\documentclass[runningheads]{llncs}

\usepackage{eccv}

\usepackage{eccvabbrv}

\usepackage{graphicx}
\usepackage{booktabs}

\usepackage[accsupp]{axessibility}  %

\usepackage{hyperref}

\usepackage{orcidlink}

\usepackage{amsmath,amsfonts,bm}
\usepackage{tabularx}
\usepackage{multirow}
\usepackage{graphicx}
\usepackage{subcaption}
\usepackage{comment}
\usepackage{makecell}
\usepackage{wrapfig}
\usepackage{soul}
\usepackage{xspace}
\usepackage{colortbl}
\usepackage[most]{tcolorbox}
\usepackage{enumitem}
\usepackage[super]{nth}
\usepackage{float}
\usepackage{pifont}

\newcommand{\bench}{ProactiveBench\xspace}

\newcolumntype{Y}{>{\centering\arraybackslash}X} %
\newcolumntype{C}[1]{>{\centering\arraybackslash}p{#1}} %
\newcolumntype{L}[1]{>{\raggedright\arraybackslash}p{#1}} %
\newcolumntype{R}[1]{>{\raggedleft\arraybackslash}p{#1}} %

\newcommand{\states}{\mathcal{S}}
\newcommand{\actions}{\mathcal{A}}
\newcommand{\policy}{\pi_\theta}
\newcommand{\rewards}{\mathcal{R}}

\newcommand{\stateo}{s_0}
\newcommand{\statet}{s_\mathit{t}}
\newcommand{\nextstatet}{s_{\mathit{t}+1}}
\newcommand{\imaget}{\mathcal{I}_\mathit{t}}
\newcommand{\prompt}{q}
\newcommand{\optionst}{\mathcal{A}_\mathit{t}}
\newcommand{\actiono}{a_0}
\newcommand{\actiont}{a_\mathit{t}}

\newcommand{\predict}{a_\mathit{c}}

\newcommand{\context}{c}

\newcommand{\llava}{LLaVA}
\newcommand{\llavacvpr}{\llava{}-1.5}
\newcommand{\llavanext}{\llava{}-NeXT}
\newcommand{\llavaov}{\llava{}-OV}
\newcommand{\qwen}{Qwen2.5-VL}
\newcommand{\smolvlm}{SmolVLM2}
\newcommand{\idefics}{Idefics3}
\newcommand{\internvl}{InternVL3}
\newcommand{\instructblip}{InstructBLIP}
\newcommand{\phimultimodal}{Phi-4-Multimodal}

\newcommand{\hint}{\textcolor{WildStrawberry}{\texttt{<hint>}}}

\definecolor{colorline}{HTML}{EDEDED}
\definecolor{reactive}{HTML}{FF9999}
\definecolor{proactive}{HTML}{66B2FF}
\definecolor{increase}{HTML}{52b69a}
\definecolor{decrease}{HTML}{dd2d4a}
\definecolor{link}{HTML}{4887cf}

\usepackage{microtype}

\renewcommand{\paragraph}[1]{\vspace{.5em}\noindent\textbf{#1}}

\begin{document}

\title{ProactiveBench: Benchmarking Proactiveness in Multimodal Large Language Models} 

\titlerunning{ProactiveBench}

\author{Thomas De Min\inst{1}
\and
Subhankar Roy\inst{2}
\and
Stéphane Lathuilière\inst{3}
\and 
Elisa Ricci\inst{1,4}
\and Massimiliano Mancini\inst{1}
}

\authorrunning{T.~De Min et al.}

\institute{ \small \begin{tabular}{c} \inst{1}University of Trento \; \inst{2}University of Bergamo  \\
\inst{3}Inria, Univ. Grenoble Alpes, CNRS, LJK  \; \inst{4}Fondazione Bruno Kessler \end{tabular} \\[0.5em] \email{thomas.demin@unitn.it} \\ \href{https://huggingface.co/datasets/tdemin16/ProactiveBench}{%
    \raisebox{-0.2ex}{\includegraphics[height=1em]{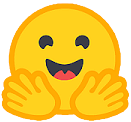}}~\textcolor{link}{\texttt{tdemin16/ProactiveBench}}%
    }\\
    \href{https://github.com/tdemin16/proactivebench}{%
    \raisebox{-0.2ex}{\includegraphics[height=1em]{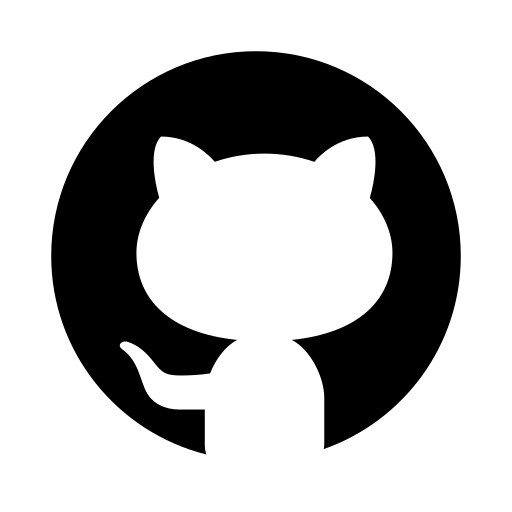}}~\textcolor{link}{\texttt{tdemin16/proactivebench}}%
}}

\maketitle

\begin{abstract}
    Effective collaboration begins with knowing when to ask for help. 
    For example, when trying to identify an occluded object, a human would ask someone to remove the obstruction. 
    Can MLLMs exhibit a similar ``proactive'' behavior by requesting simple user interventions?
    To investigate this, we introduce \bench, a benchmark built from seven repurposed datasets that tests proactiveness across different tasks such as recognizing occluded objects, enhancing image quality, and interpreting coarse sketches.  
    We evaluate 22 MLLMs on \bench, showing that (i) they generally lack proactiveness; (ii) proactiveness does not correlate with model capacity; (iii) ``hinting'' at proactiveness yields only marginal gains.
    Surprisingly, we found that conversation histories and in-context learning introduce negative biases, hindering performance.
    Finally, we explore a simple fine-tuning strategy based on reinforcement learning: its results suggest that proactiveness can be learned, even generalizing to unseen scenarios.
    We publicly release \bench as a first step toward building proactive multimodal models. 
\end{abstract}

\section{Introduction}
\label{sec:intro}
Studies in neuroscience suggest that our perception of the world arises from dynamic interaction with the environment~\cite{goodale1992separate,haskins2020active,shapiro2007embodied,heuer2020memory}. 
Faced with incomplete or ambiguous information, we instinctively generate hypotheses, proactively search for clues, and revise our interpretations. 

\begin{figure}[tp]
    \centering
    \includegraphics[width=\linewidth]{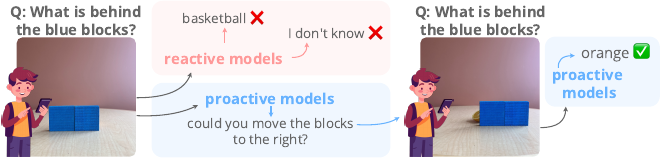}
    \caption{\textbf{Reactive \vs proactive models}. We propose \bench, the first benchmark to evaluate MLLMs' proactiveness, \ie, their ability to request additional visual cues to resolve ambiguous queries. Given an unanswerable query, a \textbf{\textcolor{reactive}{reactive}} model would either abstain or hallucinate. In contrast, a \textbf{\textcolor{proactive}{proactive}} model would ask for visual cues to disambiguate %
    the input, enabling a correct response.}
    \label{fig:teaser}
\end{figure}

This ongoing cycle of inquiry and refinement is currently unexplored for multimodal large language models (MLLMs)~\cite{zhu2025internvl3,li2024llava,bai2025qwen2}, where ambiguities may arise when a user's query is unanswerable~\cite{wu2024see,chiu2020assessing}. %
For instance, for the query \texttt{``What is behind the blue blocks?''} of \cref{fig:teaser}, a model can answer directly by hallucinating an incorrect reply~\cite{li2023evaluating}, or abstaining~\cite{whitehead2022reliable,guo2024unk}.
Such behavior is called \textit{reactive}.
Conversely, a more desirable behavior is to be \textit{proactive} and seek additional visual cues before replying.  
Yet, this is complex, as a model cannot physically act in the environment. 
However, by recalling the previous example, the user can move the blocks to reveal the hidden object. 
Currently, studies focus on reactive settings, and the proactive capabilities of MLLMs are still unknown.

To fill this gap, we study whether MLLMs \textit{can ask for help}. 
We introduce \bench{}, a novel benchmark to evaluate MLLMs' proactiveness by repurposing seven existing datasets (ROD~\cite{lee2023hardwiring}, VSOD~\cite{liao2020occlusion}, MVP-N~\cite{wang2022mvp}, ImageNet-C~\cite{hendrycks2019benchmarking}, QuickDraw~\cite{quickdraw}, ChangeIt~\cite{soucek2022lookforthechange}, and MS-COCO~\cite{lin2014microsoft}) with different target tasks (\eg, sketch recognition, product identification) that require user intervention to answer correctly.
\bench captures different aspects of proactiveness: (temporal) occlusion removal, camera movement, object movement, image quality enhancement, and asking for details. 
Each sample has a starting ambiguous frame, a reference frame with complete information, and all the frames in between. 
The user intervention, guided by the model's \textit{proactive suggestion}, produces a new frame with additional visual cues.
In total, \bench{} contains more than 108k images grouped into 18k samples featuring 19 proactive behaviors. 

We evaluate 22 state-of-the-art MLLMs (\eg, \llavaov{}~\cite{li2024llava}, \qwen{}~\cite{bai2025qwen2}, \internvl{}~\cite{zhu2025internvl3}) on \bench. %
Our experiments suggest that models lack proactiveness, either abstaining from answering or hallucinating when visual cues are insufficient (\cref{fig:teaser}). 
Using hints to elicit proactive behavior increases their %
proactive suggestion rate, but with small improvements in accuracy. 
Interestingly, while some MLLMs (\eg, \llavanext{}-Vicuna-7B, \internvl{}-1B) %
{propose more proactive suggestions} than others (\eg, \llavaov{}-7B, \qwen{}-7B, \internvl{}-8B), we show that %
this results from a lower rate of abstention on unanswerable questions, rather than a deeper understanding of the problem.
Instead, conditioning on the conversation history or few-shot samples increases %
{the proactive suggestions rate} but reduces accuracy. 
Our results highlight that proactiveness is \textit{not} an emerging property in MLLMs, showcasing the challenges of \bench.
Additionally, we show that MLLMs can learn to be proactive through post-training with GRPO~\cite{shao2024deepseekmath} equipped with tailored reward functions. 
Despite its simplicity, this approach yields substantial performance improvements over the original model and demonstrates strong generalization to unseen domains. 
While these performance are lower than those on reference images (\eg, object clearly visible, without occlusion), they suggest %
an interesting avenue for future works. %

\paragraph{Contributions:} (i) We formalize and explore MLLMs' proactiveness, promoting the development of models that can ask user assistance under uncertainty; (ii) We introduce \bench{}, an open-source benchmark to assess MLLM's proactiveness in diverse contexts; 
(iii) Our evaluation of 22 MLLMs on \bench{} reveals limited proactiveness of current models, even when explicitly hinting at being proactive, highlighting the challenges of this setting; (iv) we show that fine-tuning a model for proactiveness improves such behavior even in unseen scenarios, %
a promising direction toward building proactive MLLMs. %

\section{Related work}
\label{sec:related}

\paragraph{Benchmarking for MLLMs.}
While early efforts evaluate MLLMs on visual question answering~\cite{antol2015vqa,goyal2017making,marino2019ok}, a second wave focused on tasks requiring reasoning and world knowledge~\cite{liu2024ocrbench,li2023evaluating,liu2024mmbench,yue2024mmmu,kazemi2023geomverse}.
As recent MLLMs support multiple images and videos as inputs, more complex benchmarks have been introduced to evaluate these capabilities~\cite{kil2024compbench,kazemi2024remi,dingjie2024milebench,fu2024blink,meng2024mmiu,wang2024muirbench,tong2024eyes,jiang2024mantis,li2024mvbench}. 
Similarly, in the embodied AI literature, several studies evaluate LLMs~\cite{li2024embodied,shridhar2020alfred,padmakumar2022teach,wang2022scienceworld,savva2019habitat} integrated with agents.
However, none of these evaluate proactiveness to ambiguous or unanswerable queries.
Related to our work, \cite{wang2025actiview} and \cite{zhang2025mllms} show that MLLMs can perform complex tasks by actively seeking relevant information.
Although both assume a collaborative setting, they focus on refining predictions by exploring modifications on a single image whose query is answerable.
Liu et al.~\cite{liu2024right} explore whether MLLMs' directional guidance can support visually impaired individuals in capturing images.
However, \cite{liu2024right} limits the evaluation to a single type of proactive scenario and to single-turn conversations, not measuring the effectiveness of the MLLMs' proposed suggestions.
Instead, we investigate proactiveness in seven distinct scenarios, in which actions lead to substantial changes (\eg, viewpoints, quality, timestamp) over multiple turns for a single query. 
This enables a much more comprehensive analysis of failure cases and false proactive behaviors.

\paragraph{Active vision} improves perception~\cite{aloimonos1988active} by allowing an active observer to control sensing strategies (\eg, viewpoint) dynamically. 
Active vision has been extensively studied in view planning (\ie, determining optimal sensor viewpoints)~\cite{zeng2020view}, object recognition~\cite{browatzki2012active}, scene and 3D shape reconstruction~\cite{smith2021active}, and robotic manipulation~\cite{chuang2024active}. 
To overcome passive systems' drawbacks, \cite{xu2023active} introduces an open-world \textit{synthetic} game environment in which agents actively explore their surroundings, performing multi-round abductive reasoning.
Although we inherit the underlying spirit of active vision, our work differs in that: (i) \bench contains real-world images from diverse and complex scenarios; 
(ii) the observer receives feedback from the MLLM in natural language, fostering a collaboration of the model and the user, ideal for human-machine cooperative tasks.

\section{\bench{}}
\label{sec:method}

This section introduces \bench{}, detailing the evaluation of MLLM proactiveness (\cref{subsec:multichoice}), the {benchmark creation} (\cref{subsec:dataset}), and a filtering pipeline that ensures questions require MLLMs to ask for human intervention (\cref{subsec:filtering}). 
Model and dataset licenses are in~\cref{appx:license}.

\subsection{Evaluating proactiveness in MLLMs}
\label{subsec:multichoice}
We study MLLMs' \textit{proactiveness}, defined as the ability to either provide a correct answer or to ask for help, suggesting actions that could make the query answerable.
We evaluate proactiveness in two settings: multiple-choice question answering (MCQA) and open-ended generation (OEG).

\paragraph{MCQA evaluation.}
In this setting, models select from predefined options, allowing structured interaction with the environment and systematic assessment over multiple steps. 
We follow previous works on LLMs as agents~\cite{duan2024gtbench,liu2023agentbench} and frame the evaluation as a Markov decision process ($\states$, $\actions$, $\policy$, $\rewards$), over finite states space $\states$, discrete set of actions $\actions$, policy $\policy$ (the MLLM), and reward $\rewards$.
At step $t$, the model observes state $\statet\in\states$, comprising image $\imaget$ and valid actions $\optionst\subseteq \actions$. 
The model selects an action $\actiont$ conditioned by question $\prompt$ (\eg, \texttt{``what is this object?''}) and state $\statet=\{\imaget, \optionst\}$, \ie, $\actiont\sim\policy(\cdot\mid\prompt,\statet)$.
By selecting a proactive suggestion (\eg, \texttt{``move the occluding object''}), state $\statet$ transitions to $\nextstatet$, leading to a new image and set of valid actions.
By either abstaining (\eg, \texttt{``I do not know''}) or selecting a wrong category (\eg, dog \vs cat), the evaluation stops with a wrong prediction. 
As environments are discrete, the policy can select proactive suggestions a finite number of times, depending on the datasets, after which the evaluation terminates with a wrong prediction.
Finally, the evaluation also terminates if the model predicts the correct answer. 
Further implementation details are in the~\cref{appx:dataset_details}.

\paragraph{OEG evaluation.}
Here, the model answers queries \textit{without} predefined options. 
For this reason, evaluating OEG answers is inherently challenging as (i) they need to be interpreted and (ii) proposed actions may be inapplicable within our environments, constrained by real-world data.
Therefore, to ensure fair analyses beyond such constraints, we limit the evaluation to single-turn scenarios in OEG. 

Following prior works~\cite{liu2023visual,fu2024blink,ma2024mmlongbench,maaz2023video,song2024moviechat,nagrani2024neptune,plizzari2025omnia} we adopt an LLM-as-a-judge to score answers.
In our case, the LLM is prompted to compare the answer with both proactive suggestions and category predictions, returning a binary sequence in which each bit indicates the presence (1) or absence (0) of a \textit{valid} answer. 
A proactive suggestion is considered correct (\ie, $1$) if it is a valid mechanism to gather visual cues for the target scenario. We instruct the judge to account for variations in the answer, \eg, \texttt{``change in perspective''} is accepted for \texttt{``moving the camera''}, as implying the same outcome.
Conversely, a proactive suggestion or category is marked as absent (\ie, $0$), in the answer if it is clearly missing or not valid. %
Due to the computational cost of open-ended generation evaluation, we limit assessment to 100 examples per scenario across all scenarios of \bench{}. 
The complete LLM-as-judge prompt is provided in the \cref{appx:free_form}.

\begin{figure}[t]
    \centering
    \includegraphics[width=\linewidth]{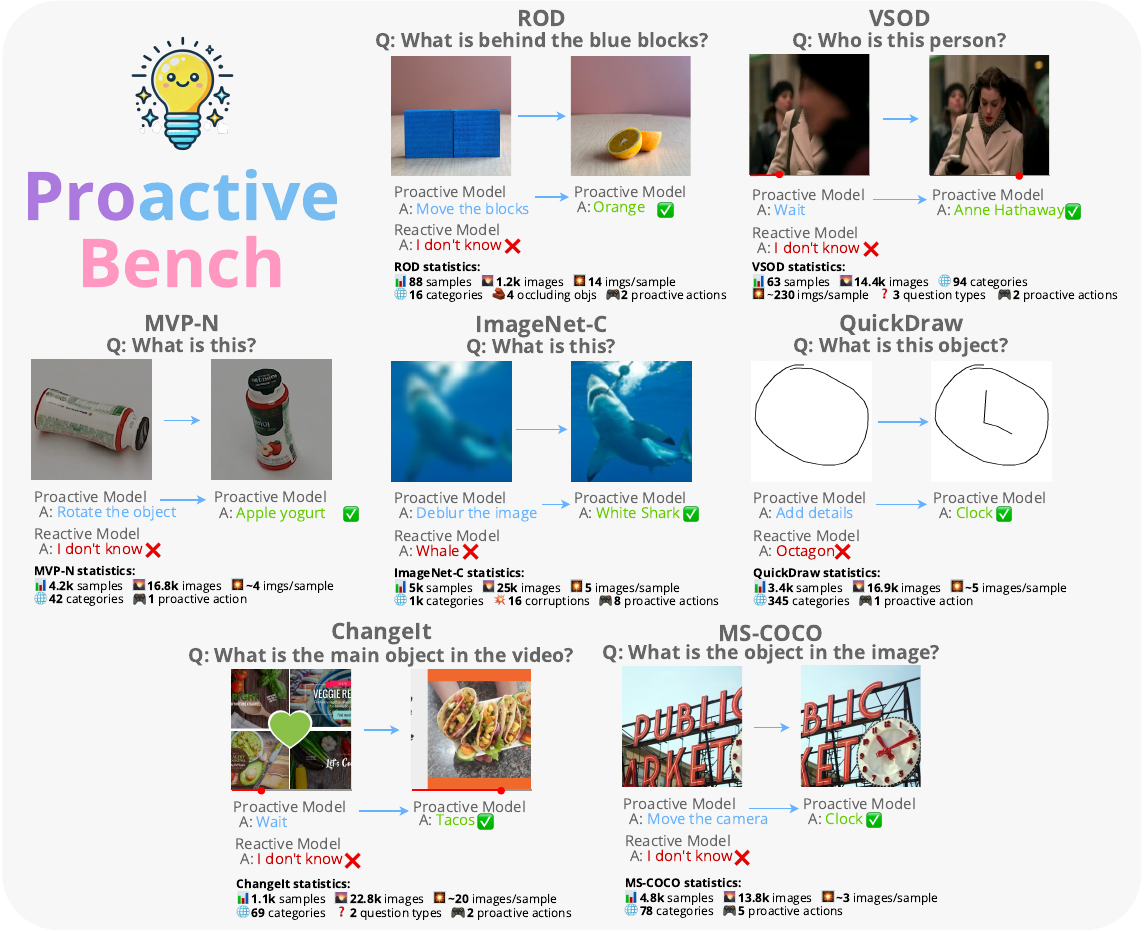}
    \caption{
    \textbf{\bench overview.} \bench evaluates proactiveness in seven scenarios. 
    The image shows examples of different scenarios and data statistics.}
    \label{fig:overview}
\end{figure}

\subsection{Benchmark construction}
\label{subsec:dataset}
We introduce seven diverse scenarios to evaluate MLLMs' proactiveness. %
We pair each scenario with a dataset that enables multi-turn interactions through proactive suggestions in the MCQA setting. 
For OEG, we expand the space of valid proactive suggestions, as it is not constrained by multi-turn evaluation. %

\paragraph{Proactive scenarios.} The proposed scenarios evaluate MLLMs' in handling:
\begin{itemize}
    \item \textbf{occluded objects} using the ROD~\cite{lee2023hardwiring} dataset, where MLLMs can ask to move the blocks to the left or right to reveal the concealed item;
    \item \textbf{temporal occlusions} with the VSOD~\cite{liao2020occlusion} dataset, suggesting to inspect frames after or before the occlusion appears;
    \item \textbf{uninformative views} via the MVP-N~\cite{wang2022mvp} dataset, proposing to rotate the object or change the camera angle to help disambiguate its semantics;
    \item \textbf{image quality improvements} using ImageNet-C (IN-C)~\cite{hendrycks2019benchmarking}, where suggesting image quality improvements reduces the uncertainty on the content;
    \item \textbf{additional visual details} through QuickDraw (QD)~\cite{quickdraw}, by asking the user for additional strokes, increasing the level of details in the drawing;
    \item \textbf{temporal ambiguities} using ChangeIt (CIT)~\cite{soucek2022lookforthechange}, where MLLMs request past or future frames to reveal the key object or action;
    \item \textbf{camera movements} with MS-COCO (COCO)~\cite{lin2014microsoft}, by asking to change the point of view (\eg, zoom, side movement) to better understand the scene.
\end{itemize}
An overview of the \bench{} scenarios is provided in \cref{fig:overview}. Additional details on each scenario are provided in \cref{appx:dataset_details}.

\paragraph{Annotation process.}
By repurposing existing datasets, we can exploit their structure and automate most of the annotation process via a rule-based procedure.
For all datasets, we use their corresponding test or validation sets.
For the large QD and IN-C, we sample $10$ and $5$ examples per category, respectively. %

A challenge in creating a proactive benchmark is modeling whether a frame is informative for the target answer. 
In this regard, ROD, MVP-N, QD, and IN-C already provide sequences ordered from least to most recognizable frames.
For example, each ROD sample has $14$ frames, with the central frame being the most occluded.
In earlier frames, the occluding object shifts left, revealing the target; in later frames, instead, it moves right.
We therefore select the least informative frame as the initial input (\eg, the first user stroke in QuickDraw). 
For CIT, we use the first video frame, which is typically uninformative for the task.
For COCO, we select images containing a single annotated bounding box and generate challenging crops of the target object (\ie, with low IoU). 
For VSOD, we manually identify frames where the target subject is fully occluded.
Category annotations are available for all datasets except VSOD. 
In this case, we annotate celebrity names if they are recognized by Google Images and discard instances where recognition fails. Full dataset details are provided in \cref{appx:dataset_details}.

Note that MLLMs may still be able to recognize the target object from the least informative frames. 
To reduce the number of cases where proactiveness is not necessary, we employ a filtering mechanism, described in the next section.

\subsection{Filtering}
\label{subsec:filtering}
As most datasets are not annotated for frame informativeness (except ROD and MVP-N), some samples (\eg, 55.3\% in ImageNet-C) can be correctly classified from the first frame (avg.\ across all MLLMs).
This allows models to bypass human intervention to cast correct predictions, leading to uneven performance across tasks. 
To focus on proactive behaviors, we filter out samples in which MLLMs can correctly guess at the first turn.
Note that this filtering step removes only samples that do not contribute to estimating proactiveness, \ie, in which the correct answer does not require multiple turns. %
Samples are filtered if they are correctly predicted at least 25\% of the time {in MCQA}, considering all {evaluated} MLLMs, \textit{during the first turn}. 
This strikes a good balance between removal and benchmark size.
After filtering, the avg.\ accuracy in the first turn drops from 32.5\% to 6.4\%, thus requiring proactive suggestions to achieve good scores.
The final benchmark counts 7,557 samples from the original size of 17,909.
We further discuss the filtering details and results on unfiltered data in \cref{appx:dataset_details}.

\section{Are MLLMs proactive?}
\label{sec:experiments}
This section evaluates multiple MLLMs using \bench{}, investigating whether they are proactive.
\Cref{subsec:experimental_setup} describes our evaluation protocol, tested models, and metrics used. 
Then, \cref{subsec:multichoice_results} describes \bench results, evaluating the proactiveness of several MLLMs.
Finally, \cref{subsec:analyses} reports additional \bench{} analysis, evaluating ways to elicit proactive suggestions.

\subsection{Experimental setup}
\label{subsec:experimental_setup}
\paragraph{Evaluation protocol.}
For each evaluation step, we feed the MLLM the question, optionally a hint to elicit proactiveness, and the current image, as \cref{subsec:multichoice} describes.
We additionally append the valid set of suggestions to the prompt for the MCQA setting, \ie, the abstain option, proactive suggestions, and four categories, only one of which is correct (see examples in \cref{appx:qualitatives}).
Hints are dataset-specific for the MCQA setting and generic for open-ended generation and lead the model towards considering proactive suggestions (\eg, \texttt{``Hint: rotating the object could provide a more informative view''} for MVP-N, and for the open-ended setting \texttt{``If you cannot answer this question, please tell me what I should do to help you''}).
The conversation history is always discarded unless explicitly mentioned (see \cref{subsec:analyses}).
Furthermore, as VSOD and ChangeIt consist of video frames, we tell the model that the visual input is taken from a video.
Finally, we rely on Qwen3-8B~\cite{yang2025qwen3} as a judge for the open-ended generation scenario, given its reliability noted by previous work~\cite{jiang2025codejudgebench}.

\paragraph{Tested models.}
We tested open and closed-weight MLLMs.
Among open-weight models we used recent and established ones: \llavacvpr{}-7B~\cite{liu2024improved}, \llavanext{}-7B~\cite{liu2024improved} with Mistral~\cite{jiang2024identifying} and Vicuna~\cite{vicuna2023} LLMs, \llavaov{}-0.5B, -7B, -72B~\cite{li2024llava}, \smolvlm{}-2.2B~\cite{marafioti2025smolvlm}, \idefics{}-8B~\cite{laurenccon2024building}, \instructblip{}~\cite{instructblip}, \qwen{}-3B, -7B, -32B, -72B~\cite{bai2025qwen2}, \internvl{}-1B, -2B, -8B, -38B, -78B~\cite{zhu2025internvl3}, \phimultimodal{}~\cite{abouelenin2025phi}.
Among closed-weight models, we considered GPT-4.1, GPT-5.2, and o4-mini \cite{openai}.

\paragraph{Metrics.}
{Since we filtered samples that leak the answer in the first turn (\cref{subsec:filtering}), we measure meaningful proactiveness via the trajectory accuracy (\textit{acc}),} \ie, the percentage %
{of proactive sequences that lead to a correct prediction}.
{Additionally, we also report} the proactive suggestions rate (\textit{psr}), %
namely the average number of human interventions requested by the model.
{For OEG,} since the evaluation is carried out over a single turn, we separately report the percentage of answers containing the correct category {(\textit{cc})} or a valid proactive suggestion (\textit{ps}).
{Note that MCQA and OEG metrics are not directly comparable, as the former measures correct answers at the end of a trajectory and the corresponding proactive suggestion rate, while the latter measures the percentage of replies that contain a correct answer or a valid proactive suggestion in the first turn.}

\subsection{MLLMs results in \bench{}}
\label{subsec:multichoice_results}

\paragraph{Multiple-choice question answering.}
\Cref{tab:zero_shot} reports MLLMs' individual performance on \bench{}.
Surprisingly, there is no %
{simple monotonic relationship between model size and proactiveness within the tested model families}, \eg, \internvl{}-1B outperforms \internvl{}-8B in accuracy ($27.1$\% \vs $12.7$\%) and proactive suggestions ($0.7$ \vs $0.3$).
Furthermore, older models (\eg, \llavacvpr{}-7B) even outperform their newer and larger counterparts (\ie, \llavaov{}-72B) by a discrete margin in \textit{acc} ($24.8$\% \vs $13.0$\%) and \textit{psr} ($0.9$ \vs $0.3$).
Interestingly, the LLM influences results, with \llavanext{} Mistral achieving lower \textit{acc} than its counterpart using Vicuna ($4.5$\% \vs $19.3$\%).
Instead, closed-source models (\eg, GPT-4.1) show the best \textit{acc}, with a low \textit{ps}.
\begin{wrapfigure}[13]{r}{0.5\linewidth}
    \centering
    \vspace{-\baselineskip}
    \includegraphics[width=\linewidth]{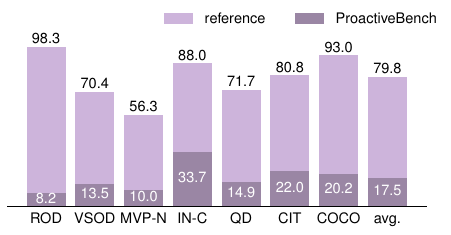}
    \caption{\textbf{Acc.\ in \bench{} \vs reference.} Models underperform by over $60$\% in scenarios that require proactiveness.}
    \label{fig:reference}
\end{wrapfigure}
Yet, they achieve extremely high accuracies on  COCO (about 3$\times$ better than other models), suggesting potential %
training data contamination. 
Unfortunately, we cannot verify this due to the proprietary nature of the data.

\begin{table*}[tp]
    \centering
    \scriptsize
    \caption{\textbf{MCQA results on \bench{}.} Accuracy (\textit{acc}) and proactive suggestion rate (\textit{psr}) of MLLMs across all \bench{} splits.}
    \resizebox{\linewidth}{!}{
    \begin{tabular}{ 
        rlcccccccccccccccc
    }
      &  & \multicolumn{2}{c}{ROD} & \multicolumn{2}{c}{VSOD} & \multicolumn{2}{c}{MVP-N} & \multicolumn{2}{c}{IN-C} & \multicolumn{2}{c}{QD} & \multicolumn{2}{c}{CIT} & \multicolumn{2}{c}{COCO} & \multicolumn{2}{c}{avg.} \\
family & model & \textit{acc} & \textit{psr} & \textit{acc} & \textit{psr} & \textit{acc} & \textit{psr} & \textit{acc} & \textit{psr} & \textit{acc} & \textit{psr} & \textit{acc} & \textit{psr} & \textit{acc} & \textit{psr} & \textit{acc} & \textit{psr}\\
\toprule
LLaVA-1.5 & 7B & 12.5 & 0.7 & 26.2 & 1.7 & 6.7 & 0.0 & 26.2 & 0.8 & 25.5 & 0.7 & \textbf{44.2} & 1.3 & 32.3 & 0.9 & 24.8 & 0.9 \\
\rowcolor{colorline} & Mistral-7B & 0.0 & 0.0 & 0.0 & 0.2 & 1.6 & 0.1 & 10.2 & 0.4 & 1.0 & 0.1 & 17.2 & 1.4 & 1.6 & 0.0 & 4.5 & 0.3 \\
\rowcolor{colorline}\multirow{-2}{*}{LLaVA-NeXT} & Vicuna-7B & 19.3 & 0.7 & 11.9 & 0.5 & 6.5 & 0.1 & 33.2 & 1.3 & 10.2 & 0.9 & 36.6 & 0.9 & 17.1 & 0.3 & 19.3 & 0.7 \\
 & 0.5B & 44.3 & 2.3 & 9.5 & 1.6 & 12.8 & 0.4 & 24.8 & 1.4 & \textbf{33.8} & 1.5 & 31.1 & 1.4 & 16.9 & 0.4 & 24.8 & 1.3 \\
 LLaVA-OV & 7B & 0.0 & 0.0 & 14.3 & 0.4 & 6.7 & 0.0 & 27.8 & 1.0 & 24.3 & 0.4 & 10.4 & 0.3 & 3.2 & 0.0 & 12.4 & 0.3 \\
 & 72B & 0.0 & 0.0 & 19.0 & 0.4 & 5.0 & 0.1 & 32.2 & 1.2 & 14.3 & 0.2 & 16.9 & 0.5 & 3.7 & 0.0 & 13.0 & 0.3 \\
\rowcolor{colorline}SmolVLM2 & 2.2B & 0.0 & 0.0 & 11.9 & 0.2 & 11.1 & 0.1 & 19.5 & 1.0 & 9.9 & 0.6 & 25.5 & 0.6 & 5.8 & 0.0 & 12.0 & 0.4 \\
Idefics3 & 8B & 31.8 & 1.6 & 19.0 & 2.2 & 7.4 & 0.1 & 32.1 & 1.1 & 12.5 & 0.6 & 12.1 & 0.4 & 9.0 & 0.2 & 17.7 & 0.9 \\
\rowcolor{colorline}InstructBLIP & 7B & 0.0 & 0.0 & 9.5 & 1.3 & 8.8 & 0.1 & 11.3 & 0.0 & 18.3 & 0.1 & 24.5 & 0.0 & 12.6 & 0.0 & 12.2 & 0.2 \\
 & 3B & 0.0 & 0.0 & 9.5 & 0.0 & 4.9 & 0.0 & 35.9 & 2.0 & 7.9 & 0.2 & 12.4 & 0.3 & 6.3 & 0.0 & 11.0 & 0.4 \\
 & 7B & 0.0 & 0.0 & 0.0 & 0.0 & 4.3 & 0.0 & 40.5 & 1.3 & 9.9 & 0.1 & 9.8 & 0.1 & 4.9 & 0.0 & 9.9 & 0.2 \\
 & 32B & 0.0 & 0.0 & 4.8 & 0.0 & 4.6 & 0.0 & 30.9 & 0.4 & 12.3 & 0.0 & 17.4 & 0.4 & 5.5 & 0.0 & 10.8 & 0.1 \\
\multirow{-4}{*}{Qwen-2.5-VL} & 72B & 0.0 & 0.0 & 2.4 & 0.2 & 6.7 & 0.0 & 29.2 & 0.9 & 3.1 & 0.1 & 9.3 & 0.3 & 2.0 & 0.0 & 7.5 & 0.2 \\
\rowcolor{colorline} & 1B & \textbf{61.4} & 2.1 & 21.4 & 0.3 & 19.7 & 0.4 & 38.6 & 1.1 & 15.0 & 0.5 & 16.9 & 0.3 & 16.5 & 0.1 & 27.1 & 0.7 \\
\rowcolor{colorline} & 2B & 1.1 & 0.0 & \textbf{31.0} & 0.3 & \textbf{20.1} & 0.2 & 46.1 & 1.5 & 18.1 & 0.5 & 28.5 & 0.6 & 29.7 & 0.2 & 24.9 & 0.5 \\
\rowcolor{colorline}InternVL3 & 8B & 0.0 & 0.0 & 11.9 & 0.2 & 6.4 & 0.0 & 37.7 & 1.0 & 15.4 & 0.5 & 10.1 & 0.2 & 7.1 & 0.0 & 12.7 & 0.3 \\
\rowcolor{colorline} & 38B & 0.0 & 0.0 & \textbf{31.0} & 2.3 & 12.5 & 0.2 & 45.5 & 0.7 & 16.8 & 0.5 & 27.0 & 1.0 & 28.4 & 0.2 & 23.0 & 0.7 \\
\rowcolor{colorline} & 78B & 0.0 & 0.0 & 16.7 & 0.3 & 10.7 & 0.0 & 39.8 & 0.1 & 5.3 & 0.0 & 17.4 & 0.4 & 19.2 & 0.0 & 15.6 & 0.1 \\
Phi-4-Multimodal & 6B & 1.1 & 0.0 & 16.7 & 1.0 & 18.9 & 0.0 & 29.8 & 1.6 & 21.9 & 0.4 & 32.6 & 0.6 & 15.2 & 0.2 & 19.4 & 0.5 \\
\rowcolor{colorline} & o4-mini & 0.0 & 0.0 & 16.7 & 0.6 & 19.8 & 0.0 & 49.0 & 0.2 & 21.6 & 0.0 & 37.9 & 0.8 & 92.8 & 0.0 & \textbf{34.0} & 0.2 \\
\rowcolor{colorline} & GPT-4.1 & 0.0 & 0.0 & 0.0 & 0.2 & 15.2 & 0.1 & \textbf{68.2} & 1.1 & 15.0 & 0.2 & 23.5 & 0.6 & \textbf{94.4} & 0.0 & 30.9 & 0.3 \\
\rowcolor{colorline}\multirow{-3}{*}{OpenAI} & GPT-5.2 & 0.0 & 0.0 & 0.0 & 0.2 & 7.8 & 0.1 & 36.6 & 0.3 & 13.6 & 0.1 & 21.7 & 0.5 & 93.6 & 0.0 & 24.8 & 0.2 \\

    \end{tabular}
    }
    \label{tab:zero_shot}
\end{table*}

To put these results in perspective, \cref{fig:reference} compares accuracy (avg.\ over all models) in \bench{} with the \textit{reference} setting, where we directly prompt MLLMs with the reference frame (\ie, with no occlusions/ambiguity).
The goal is to disentangle the recognition ability of MLLMs and their proactiveness. 
While MLLMs correctly classify $79.8$\% of samples in the reference setting, they underperform by more than $60$\% when tasked with navigating to the correct answer through proactive suggestions. 
The discrepancy is quite stark in the ROD dataset, where models achieve $8.2$\% \textit{acc}, while the reference counterpart reaches $98.3$\% on average.
This demonstrates a severe lack of MLLMs' proactiveness.

\begin{figure}[t]
    \includegraphics[width=\linewidth]{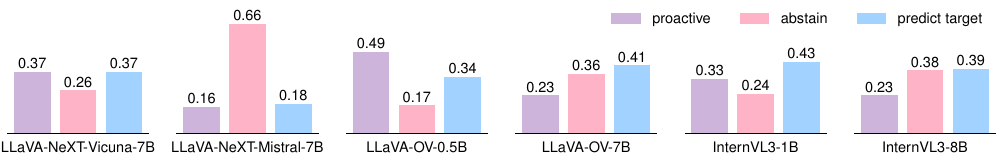}
    \caption{\textbf{Action distributions.} While \llavaov{}-7B, \internvl{}-8B, and \llavanext{}-Mistral-7B abstain or guess an answer, the other models prioritize proactive suggestions; thus, leveraging better visual cues and making better predictions.}
    \label{fig:dist_zero_shot}
\end{figure}

We further investigate proactiveness by visualizing the action distributions, averaged across all scenarios, for proactive, abstain, and target category predictions in \cref{fig:dist_zero_shot}. 
Specifically, we compare pairs of MLLMs having different LLMs (\ie, \llavanext{} Mistral and Vicuna) and different parameter counts (\ie, \llavaov{}-0.5B and -7B, \internvl{}-1B and -8B). 
While \llavaov{}-7B, \internvl{}-8B, and \llavanext{} Mistral tend to abstain over sampling proactive suggestions (likely due to different training data and/or model sizes), the other three show the exact opposite behavior.
Thus, they are more likely to {be proactive} %
(over $2\times$ as likely for \llavaov{}-0.5B) and, as a result, reach higher accuracy.
A similar behavior was reported in~\cite{wolfe2024laboratory}, with \llavanext{} Mistral abstaining more than \llavanext{} Vicuna.
Further results are in the \cref{appx:extended_results}.

\begin{table*}[t]
    \centering
    \scriptsize
    \caption{{\textbf{OEG results on \bench{}.} Percentage of correct category predictions (\textit{cc}) and valid proactive suggestions (\textit{ps}) of MLLMs across all \bench{} splits.} %
    }
    \resizebox{\linewidth}{!}{
    \begin{tabular} { 
        rlcccccccccccccccc
    }
     &  & \multicolumn{2}{c}{ROD} & \multicolumn{2}{c}{VSOD} & \multicolumn{2}{c}{MVP-N} & \multicolumn{2}{c}{IN-C} & \multicolumn{2}{c}{QD} & \multicolumn{2}{c}{CIT} & \multicolumn{2}{c}{COCO} & \multicolumn{2}{c}{avg.} \\
    family & model & \textit{cc} & \textit{ps} & \textit{cc} & \textit{ps} & \textit{cc} & \textit{ps} & \textit{cc} & \textit{ps} & \textit{cc} & \textit{ps} & \textit{cc} & \textit{ps} & \textit{cc} & \textit{ps} & \textit{cc} & \textit{ps} \\
\toprule
 LLaVA-1.5 & 7B & 0.0 & 5.7 & 0.0 & 0.0 & 0.0 & 0.0 & 1.0 & 2.0 & 1.0 & 6.0 & \textbf{3.0} & 0.0 & 1.0 & 1.0 & 0.9 & 2.1 \\
 \rowcolor{colorline} & Mistral-7B & 0.0 & 3.4 & 0.0 & \textbf{2.4} & 0.0 & 0.0 & 1.0 & 34.0 & \textbf{4.0} & \textbf{27.0} & 2.0 & 6.0 & 4.0 & 1.0 & 1.6 & 10.5 \\
 \rowcolor{colorline}\multirow{-2}{*}{LLaVA-NeXT} & Vicuna-7B & 0.0 & 5.7 & 2.4 & 0.0 & 0.0 & 0.0 & 3.0 & \textbf{36.0} & 0.0 & 23.0 & \textbf{3.0} & 6.0 & 0.0 & \textbf{4.0} & 1.2 & \textbf{10.7} \\
  & 0.5B & 0.0 & 1.1 & 0.0 & 0.0 & 0.0 & 0.0 & 0.0 & 11.0 & 2.0 & 2.0 & 1.0 & 0.0 & 0.0 & 1.0 & 0.4 & 2.2 \\
  LLaVA-OV & 7B & 0.0 & 1.1 & 0.0 & \textbf{2.4} & 0.0 & 0.0 & 2.0 & 18.0 & 0.0 & 9.0 & 0.0 & 2.0 & 2.0 & 2.0 & 0.6 & 4.9 \\
  & 72B & 0.0 & 5.7 & 0.0 & 0.0 & 0.0 & 0.0 & 2.0 & 19.0 & 2.0 & 7.0 & 0.0 & 1.0 & 0.0 & 1.0 & 0.6 & 4.8 \\
 \rowcolor{colorline}SmolVLM2 & 2.2B & 0.0 & 0.0 & 0.0 & 0.0 & 0.0 & 0.0 & 0.0 & 5.0 & 3.0 & 0.0 & 0.0 & 0.0 & 0.0 & 1.0 & 0.4 & 0.9 \\
 Idefics3 & 8B & 0.0 & 1.1 & 2.4 & 0.0 & 0.0 & 0.0 & 2.0 & 7.0 & 0.0 & 1.0 & 0.0 & 0.0 & 4.0 & 1.0 & 1.2 & 1.4 \\
 \rowcolor{colorline} & 3B & 0.0 & 1.1 & 2.4 & 0.0 & 0.0 & 0.0 & 2.0 & 22.0 & 2.0 & 3.0 & 0.0 & 1.0 & 3.0 & 0.0 & 1.3 & 3.9 \\
 \rowcolor{colorline} & 7B & 0.0 & \textbf{6.8} & 2.4 & 0.0 & 0.0 & \textbf{1.0} & 3.0 & 31.0 & 0.0 & 16.0 & 1.0 & 8.0 & 2.0 & 3.0 & 1.2 & 9.4 \\
 \rowcolor{colorline} & 32B & 0.0 & 3.4 & 0.0 & 0.0 & 0.0 & 0.0 & 2.0 & 25.0 & 0.0 & 3.0 & 2.0 & \textbf{9.0} & 0.0 & 0.0 & 0.6 & 5.8 \\
 \rowcolor{colorline}\multirow{-4}{*}{Qwen-2.5-VL} & 72B & 0.0 & 2.3 & 0.0 & 0.0 & 0.0 & 0.0 & 3.0 & 28.0 & 1.0 & 11.0 & 1.0 & 6.0 & 4.0 & 2.0 & 1.3 & 7.0 \\
  & 1B & 0.0 & 1.1 & 2.4 & 0.0 & 0.0 & 0.0 & 3.0 & 17.0 & 2.0 & 5.0 & 1.0 & 1.0 & 3.0 & 3.0 & 1.6 & 3.9 \\
  & 2B & 0.0 & 0.0 & 2.4 & 0.0 & 0.0 & 0.0 & 2.0 & 17.0 & 0.0 & 3.0 & 1.0 & 1.0 & 0.0 & 0.0 & 0.8 & 3.0 \\
 InternVL3 & 8B & 0.0 & 1.1 & \textbf{4.8} & 0.0 & 0.0 & 0.0 & 3.0 & 18.0 & 0.0 & 1.0 & 1.0 & 2.0 & 2.0 & 0.0 & 1.5 & 3.2 \\
  & 38B & 0.0 & 3.4 & \textbf{4.8} & 0.0 & 0.0 & 0.0 & 8.0 & 15.0 & 1.0 & 1.0 & 1.0 & 6.0 & 4.0 & 0.0 & 2.7 & 3.6 \\
  & 78B & 0.0 & 1.1 & 2.4 & 0.0 & 0.0 & 0.0 & \textbf{11.0} & 16.0 & 1.0 & 5.0 & \textbf{3.0} & 3.0 & \textbf{7.0} & 0.0 & \textbf{3.5} & 3.6 \\
    \end{tabular}
    }
    \label{tab:free_form}
\end{table*}

\paragraph{Open-ended generation.}
\Cref{tab:free_form} reports MLLMs' %
{percentage of correct first-turn predicted categories (\textit{cc}) and valid proactive suggestions (\textit{ps})} in OEG.
Overall, even when models are not restricted to multiple-choice options, they still fail to be proactive; instead, they either abstain or hallucinate answers, much like in the MCQA setting.
Similarly, there is no %
{simple monotonic relationship between model size and proactiveness within the tested model families}, suggesting that proactiveness is not a property that emerges with scale.
Surprisingly, by allowing \llavanext{}-Mistral to answer without constraints, it overcomes the issue with the abstention rate of \cref{tab:zero_shot}, showing the {second best accuracy in predicting valid proactive suggestions.}
On the other hand, \internvl{}-1B superiority among open-weight models in the MCQA setting is not observed in the OEG scenario. %

{By examining \textit{cc} and \textit{ps}, we observe that the model is more accurate when proposing proactive suggestions than predicting correct categories, although their overall accuracy remains low.} 
This behavior can be attributed to three main factors. 
{First}, we instruct the LLM-as-judge to allow for flexible rephrasings to match MLLMs' answers to the finite set of available actions, broadening the range of valid suggestions.
{Second}, removing possible categories from the prompt increases model uncertainty, lowering categorization accuracy. 
{Third}, filtering (\cref{subsec:filtering}) makes the single-turn setting more challenging, as the model cannot rely on additional visual cues. 

Overall, OEG performances are worse than in the MCQA setting.
{By mapping every inapplicable proactive suggestion to a random valid one, we can extend OEG to multiple turns and perform an approximate comparison between the two settings under the same metrics.}
The worst model in MCQA (\llavanext{}-Mistral-7B) achieves %
$4.5$\% in \textit{acc} with $0.3$ \textit{psr}. %
Instead, the best evaluated MLLM in {multi-turn} OEG (\qwen{}-7B) only achieves %
{$3.9$\% \textit{acc} with $0.1$ \textit{psr}}. %
This confirms that MLLMs are still far from being proactive.
Full results are in \cref{appx:extended_results}.

\begin{figure}[tp]
    \includegraphics[width=\linewidth]{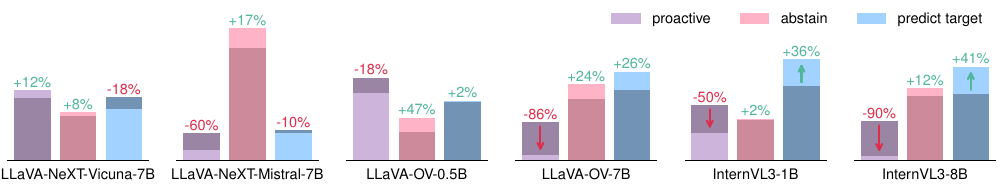}
    \caption{\textbf{Action distributions with random proactive options.} Lighter bars describe variations when replacing valid proactive suggestions with invalid ones. We color-code \textcolor{increase}{positive} and \textcolor{decrease}{negative} changes in action prob. If models still assign high prob.\ with random proactive actions, it implies they are not proactive and just avoid abstention.}
    \label{fig:dist_random}
\end{figure}
\subsection{Analyzing and eliciting MLLMs proactiveness}
We now analyze MLLMs' proactiveness, investigating the influence of different prompting strategies. %
We focus these analyses on MCQA for its higher controllability and %
multi-turn evaluation, 
extending to OEG where possible.

\label{subsec:analyses}
\paragraph{Why some MLLMs {are more likely to propose proactive suggestions}?}
To answer this question, we replaced valid proactive suggestions with invalid ones chosen randomly from other datasets (\eg, \texttt{``rewind the video''} for QuickDraw).
If models that %
{propose proactive suggestions more frequently} still choose (invalid) proactive options, this implies that they are %
{just} guessing (even incorrectly) over abstaining.
We limit this evaluation to the MCQA setting, as it allows for a more controlled examination.
\Cref{fig:dist_random} shows the action distribution with the same six models as \cref{fig:dist_zero_shot}, averaged for across all datasets.
Replacing valid proactive suggestions with invalid ones substantially reduces %
{proactive suggestion rate} for \llavanext{} Mistral, \llavaov{}-7B, and \internvl{}-8B (\ie, $-60$\%, $-86$\%, and $-90$\% relative decrease, respectively). 
Instead, models that %
{predict more proactive suggestions} in \cref{fig:dist_zero_shot}, still select proactive options in \cref{fig:dist_random}, even if the latter are now random and not applicable to the input scenario. 
\llavanext{} Vicuna even increases the probability of sampling proactive suggestions (from $37$\% to $49$\%).
These insights indicate that models showing a higher rate of proactive suggestions are \textbf{not} necessarily more proactive, but rather they are less prone to abstain~\cite{shukor2023beyond}, preferring unknown answers.
{Indeed, meaningful proactiveness occurs if and only if a model uses proactive suggestions to reach better states and improve answer accuracy.}
Full results are in \cref{appx:extended_results}.

\begin{figure*}[t]
    \centering
    \begin{subfigure}[b]{0.32\linewidth}
        \includegraphics[width=\linewidth]{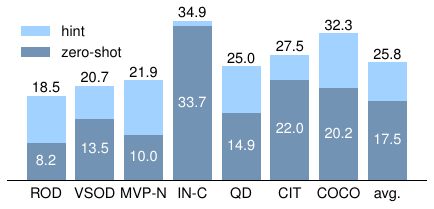}
        \caption{MCQA \textit{acc}.}
        \label{fig:hint_acc}
    \end{subfigure}
    \hfill
    \begin{subfigure}[b]{0.32\linewidth}
        \includegraphics[width=\linewidth]{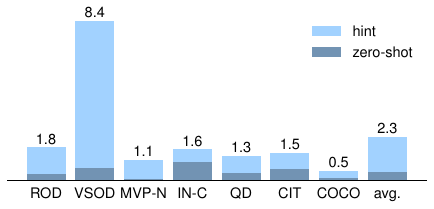}
        \caption{MCQA \textit{psr}.}
        \label{fig:hint_turn}
    \end{subfigure}
    \hfill
    \begin{subfigure}[b]{0.32\linewidth}
        \includegraphics[width=\linewidth]{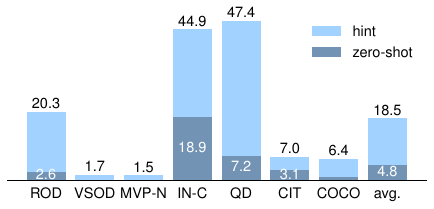}
        \caption{Open-ended gen.\ \textit{ps}.}
        \label{fig:hint_open-ended}
    \end{subfigure}%
    \caption{\textbf{Conditioning models with hints} for MCQAs and OEG. Results are averaged across all MLLMs. Zero-shot refers to models not prompted with hints.}
    \label{fig:hint}
\end{figure*}
\begin{figure}[t]
    \includegraphics[width=\linewidth]{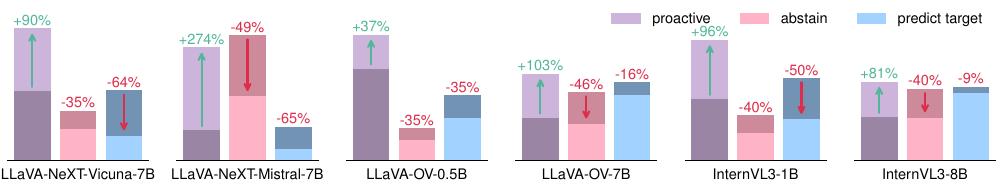}
    \caption{\textbf{Action distributions with hints.} Bars describe action distributions with (light) or without (dark) hints in the prompt. We color-code \textcolor{increase}{positive} and \textcolor{decrease}{negative} changes in action probabilities. Hinting increases the prob.\ of proactive suggestions.}
    \label{fig:dist_hint}
\end{figure}

\paragraph{Does hinting boost proactiveness?}
Explicitly hinting at proactive suggestions may elicit MLLMs' proactiveness, helping to navigate to the correct answer.
To evaluate this hypothesis, we add \textit{dataset-specific} hints to MCQA prompts (\eg, for ROD \texttt{``Hint: moving the occluding object might reveal what is behind it''}) and \textit{generic} ones for the OEG (\ie, \texttt{``If you cannot answer this question, please tell me what I should do to help you''}).
We report the full list of hints used in \cref{appx:dataset_details,appx:free_form}. Results are shown in \cref{fig:hint}, in terms of MCQA and OEG metrics. %
Extended results are in \cref{appx:extended_results}.

\Cref{fig:hint_turn} shows that hinting increases the \textit{psr} in MCQA by $1.9$ on average, with a significant boost in VSOD, likely due to its large exploration space.
Nonetheless, the accuracy (\cref{fig:hint_acc}) does not surpass the random choice on average, reaching $25.8$\% ($+8.3$\%) and suggesting that hinting cannot elicit proactiveness from current MLLMs.
Additionally, in $16.0$\% of cases, MLLMs blindly choose proactive suggestions, disregarding the original task and reaching the maximum exploration steps allowed by the environment, failing to predict the correct category.
Hinting also increases %
{the percentage of valid proactive suggestions (\textit{ps})} in OEG (\cref{fig:hint_open-ended}), with IN-C and QD showing the largest gains. 
Unlike other tasks that require a deeper understanding of the concept (\eg, rotating the camera for MVP-N), IN-C and QD require the model to request image-quality improvements and additional details about the drawing, %
likely easier to interpret. 

Although hinting promotes %
{predicting proactive suggestions}, models may over-exploit proactive suggestions, failing the classification even if they stumble across the reference image.
\Cref{fig:dist_hint} further visualizes this by showing how action distributions change w.r.t. the six models in \cref{fig:dist_zero_shot}.
While original distributions (darker colors) suggest that models infrequently choose proactive options, hints completely changes this behavior, with models preferring hinted actions over correct predictions{, confirming their lack of proactiveness}.

\begin{figure*}[tp]
    \centering
    \begin{subfigure}[b]{0.49\linewidth}
        \includegraphics[width=\linewidth]{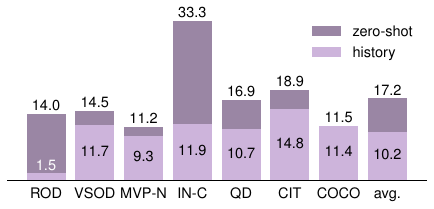}
        \caption{Avg.\ accuracy per dataset.}
        \label{fig:history_acc}
    \end{subfigure}
    \hfill
    \begin{subfigure}[b]{0.49\linewidth}
        \includegraphics[width=\linewidth]{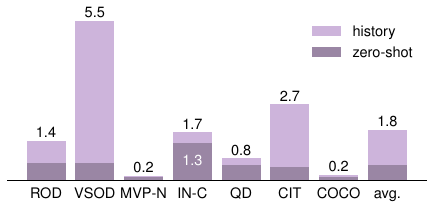}
        \caption{Avg.\ proactive suggestions per dataset.}
        \label{fig:history_turn}
    \end{subfigure}%
    \caption{\textbf{Conditioning on conversation histories.} Results are averaged across all MLLMs that support multi-image inference. Zero-shot refers to models not conditioned on histories. Conversation histories increase proactiveness but hurt accuracy.
    }
    \label{fig:history}
\end{figure*}
\paragraph{Does knowledge of the past elicit proactiveness?}
While MLLMs observe only the current state (\cref{subsec:multichoice}), a key question is whether conditioning models on previous states and actions (conversation history) elicits proactiveness, \ie, $\policy(\cdot\mid\prompt, \stateo, \actiono, ..., \statet)$.
\Cref{fig:history} shows that the average \textit{acc} drops by $7$\% while the \textit{psr} increases from $0.5$ to $1.8$ on average, compared to the zero-shot case{, suggesting that proactiveness cannot be elicited simply through conversation histories}.
Although models are not explicitly ``told'' to be proactive, like in \cref{fig:hint}, past proactive suggestions bias models towards repeating them.
MLLMs exhibited the same behavior as with ``hints'', repeatedly selecting proactive suggestions until reaching the maximum number of allowed steps, occurring in $12.9$\% of the cases.
This is lower than the $16.0$\% observed with hints because the first action is always unconditioned; therefore, blind selection of proactive actions occurs only if the first action is also proactive, biasing subsequent substeps.

\paragraph{Do few-shots improve proactiveness?}
We now investigate whether conditioning the policy on a few correct examples elicits proactiveness.
Let $\context=(\prompt^c,\stateo^c, \actiono^c, ..., \statet^c, \predict^c)$ be a conversation example leading to the correct answer $\predict^c$. 
We condition the action sampling on $m$ of such examples, $\policy(\cdot\mid\context_0, ..., \context_\mathit{m}, \prompt, \statet)$ on ROD and MVP-N.
Compared to other datasets, these two provide dense annotations indicating which frames contain a recognizable instance of the target object. 
This enables the automatic generation of few-shot samples composed of action sequences that transition from an ambiguous to a predictable state through proactive suggestions.
We experiment with $m=1$ and $m=3$.
\begin{wrapfigure}[12]{r}{0.5\linewidth}
    \centering
    \vspace{-\baselineskip}
    \begin{subfigure}{0.49\linewidth}
        \includegraphics[width=\linewidth]{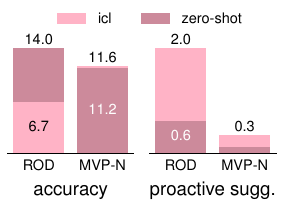}
        \caption{$1$ sample}
        \label{fig:icl_self_1}
    \end{subfigure}%
    \hfill
    \begin{subfigure}{0.49\linewidth}
        \includegraphics[width=\linewidth]{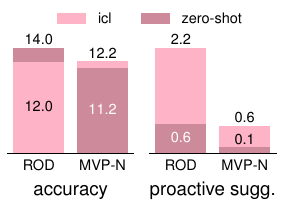}
        \caption{$3$ samples}
        \label{fig:icl_self_3}
    \end{subfigure}
    \caption{\textbf{Conditioning on few shots.} Results are averaged across all MLLMs. Zero-shot refers to models not using ICL.}
    \label{fig:icl_self}
\end{wrapfigure}
\Cref{fig:icl_self} shows how proactiveness changes with few-shot in-context learning (ICL).
Compared to the base setting (zero-shot), the avg.\ \textit{psr} increases by $1.4$ and $0.2$ on ROD and MVP-N, and $1.6$ and $0.5$ with one and three samples, respectively.
The accuracy drops in ROD while remaining stable in MVP-N, resulting in $6.7$\% and $11.6$\% with one sample, and $12.0$\% and $12.2$\% with three.
When conditioning with one sample on ROD, models either tend to predict the same category of the ICL example or blindly select proactive suggestions until reaching the maximum number of exploration steps.
Scaling ICL to three examples increases \textit{acc} in some models (\eg,\llavaov{} 7B and \phimultimodal{}).
Generally, \internvl{}-1B and \llavaov{}-0.5B are the most prone to repeating proactive suggestions and disregarding the main task, while \internvl{}-8B and \smolvlm{}-2.2B tend to abstain.
Similarly, in MVP-N, model errors arise either from random guesses, abstentions, or, occasionally, %
proactive sequences ending with incorrect predictions.

\section{Can MLLMs learn proactiveness from data?}
\label{sec:learning_proactiveness}
In the previous section, we investigated whether proactive behavior could be elicited from MLLMs using specific conditioning strategies, but observed only marginal improvements.
We now test whether proactiveness can be effectively learned and if models fine-tuned for proactiveness generalize to unseen scenarios.

\paragraph{Training for proactiveness.}
To build a training dataset that enables learning proactiveness, we follow a procedure similar to that described in \cref{sec:method} for constructing the MCQA setting.
We train models using two scenarios, QuickDraw and COCO, because (i) they provide sufficient training data, (ii) they cover both abstract and natural images, and (iii) restricting training to a subset of datasets allows us to evaluate generalization to unseen scenarios.
For COCO, we use its corresponding training split, while for QuickDraw, we sampled a subset that is disjoint from that used in \bench{}.
To reduce the computational requirements and simplify optimization, we limit the training set to single-turn interactions, sampling both ambiguous and unambiguous frames. 
These allow the model to learn which situations require proactiveness and which direct prediction.

Knowing when to propose proactive suggestions and when to predict the correct answer requires dense annotations that we do not have and that are generally not available in standard scenarios. 
To overcome the need for such annotations, we train MLLMs using reinforcement learning (RL), rewarding models to jointly prioritize response efficiency (low \textit{psr}) and \textit{acc}.
We use Group-Relative Policy Optimization (GRPO)~\cite{shao2024deepseekmath}, sampling $8$ answers for each prompt-image pair, and rewarding each answer via the following rule: $r_c=1$, if the answer corresponds to the correct category, $r_p\in\{0.5, 0.75, 1.0\}$, if it is a valid proactive suggestion, and $r_w=0$ otherwise.
Intuitively, if the reward for proactive suggestions is lower than correct category predictions, the model should learn to prioritize class predictions and revert to proactive suggestions when uncertain about the correct answer.
Further details are in \cref{appx:hyperparams}.

\begin{table}[tp]
    \centering
    \scriptsize
    \caption{\textbf{Learning proactiveness.} \llavanext{}-Mistral-7B and \qwen{}-3B \textit{acc} and \textit{psr} with RL post-training. We report the original model, RL post-training with proactive reward $r_p\in\{0.5, 0.75, 1.0\}$, and the reference performance.}
    \resizebox{\linewidth}{!}{
    \begin{tabular}{llcccccccccccccccc}
          && \multicolumn{4}{c}{\textit{in-domain}} & \multicolumn{10}{c}{\textit{out-of-domain}} & \\
          \cmidrule(lr){3-6}
          \cmidrule(lr){7-16}
          && \multicolumn{2}{c}{QD} & \multicolumn{2}{c}{COCO} & \multicolumn{2}{c}{ROD} & \multicolumn{2}{c}{VSOD} & \multicolumn{2}{c}{MVP-N} & \multicolumn{2}{c}{IN-C} & \multicolumn{2}{c}{CIT} & \multicolumn{2}{c}{avg.} \\
        model & config.\ & \textit{acc} & \textit{psr} & \textit{acc} & \textit{psr} & \textit{acc} & \textit{psr} & \textit{acc} & \textit{psr} & \textit{acc} & \textit{psr} & \textit{acc} & \textit{psr} & \textit{acc} & \textit{psr} & \textit{acc} & \textit{psr}\\
        \toprule
         & original & 1.0 & 0.1 & 1.6 & 0.0 & 0.0 & 0.0 & 0.0 & 0.2 & 1.6 & 0.1 & 10.2 & 0.4 & 17.2 & 1.4 & 4.5 & 0.3 \\
         & $r_p = 0.5$ & \textbf{43.6} & 1.1 & 57.3 & 1.0 & \textbf{36.4} & 0.9 & 26.2 & 1.7 & 13.7 & 0.2 & 58.6 & 0.6 & 47.2 & 2.2 & 40.4 & 1.1 \\
         & $r_p = 0.75$ & 42.6 & 1.1 & 56.6 & 1.0 & 34.1 & 0.8 & 26.2 & 1.7 & 14.1 & 0.2 & \textbf{59.8} & 0.7 & 42.7 & 2.1 & 39.4 & 1.1 \\
         & $r_p = 1.0$ & 11.3 & 4.8 & \textbf{79.0} & 2.6 & 19.3 & 1.1 & \textbf{69.0} & 9.7 & \textbf{27.1} & 1.1 & 20.9 & 2.8 & \textbf{58.1} & 4.3 & \textbf{40.7} & 3.7 \\
        \multirow{-5}{*}{\makecell[l]{\llavanext{}\\Mistral-7B}} & \textit{reference} & 65.6 & - & 95.5 & - & 100.0 & - & 57.1 & - & 43.6 & - & 88.6 & - & 75.3 & - & 75.1 & - \\ [2pt]
        \rowcolor{colorline} & original & 7.9 & 0.2 & 6.3 & 0.0 & 0.0 & 0.0 & 9.5 & 0.0 & 4.9 & 0.0 & 35.9 & 2.0 & 12.4 & 0.3 & 11.0 & 0.4 \\
        \rowcolor{colorline} & $r_p = 0.5$& 42.9 & 1.2 & 46.1 & 0.6 & 11.4 & 0.1 & \textbf{38.1} & 0.7 & \textbf{18.3} & 0.1 & \textbf{52.5} & 2.2 & 52.8 & 1.5 & 37.4 & 0.9 \\
        \rowcolor{colorline} & $r_p = 0.75$ & \textbf{46.2} & 2.0 & 49.4 & 0.8 & 13.6 & 0.2 & \textbf{38.1} & 0.9 & 17.4 & 0.2 & 50.1 & 2.4 & \textbf{55.6} & 1.7 & \textbf{38.6} & 1.2 \\
        \rowcolor{colorline} & $r_p = 1.0$ & 0.0 & 5.0 & \textbf{91.3} & 5.2 & \textbf{14.8} & 11.6 & 2.4 & 60.0 & 0.0 & 3.0 & 5.1 & 2.1 & 6.8 & 18.6 & 17.2 & 15.1 \\
        \rowcolor{colorline}\multirow{-5}{*}{\qwen{}-3B} & \textit{reference} & 65.5 & - & 96.0 & - & 100.0 & - & 78.6 & - & 51.7 & - & 91.5 & - & 84.8 & - & 81.2 & - \\
    \end{tabular}
    }
    \label{tab:ft}
\end{table}

\paragraph{Results.}
We compare proactiveness in MCQA pre- and post-RL for \llavanext{}-Mistral-7B, the worst-performing model in the MCQA setting (see \cref{tab:zero_shot}), and for \qwen{}-3B, one of the most widely used MLLMs.
\Cref{tab:ft} compares MLLMs tuned with different $r_p$ values with their original counterpart and the \textit{reference} setting (using reference frames as in \cref{fig:reference}).
Both models outperform all previously evaluated MLLMs in \cref{tab:zero_shot} ($37.4$\% \vs $34.0$\% of o4-mini), except for \qwen{}-3B with $r_p = 1.0$.
To this extent, setting the proactive suggestions reward lower than correct predictions, $r_p < r_c$, generally strikes a good balance between effectiveness (\eg, $37.4$\% and $38.6$\% in \textit{acc}) and efficiency ($0.9$ and $1.2$ in \textit{psr}).
Indeed, proactive suggestions rate increases as $r_p$ grows, \eg, from $0.4$ to $15.1$ \textit{psr} for \qwen{}-3B.
By setting $r_p = r_c$, \qwen{}-3B excessively generates proactive suggestions, rarely producing correct predictions, lowering accuracy.
Notably, we witness consistent behaviors across both seen and unseen scenarios, showing that proactiveness, once learned, generalizes to unseen domains.
For instance, CIT accuracy grows from $12.4$\% to $55.6$\% after post-training \qwen{}-3B with $r_p=0.75$, increasing \textit{psr} from $0.3$ to $1.7$, while the accuracy decreases to $5.4$\% when $r_p=1.0$, reaching \textit{psr} of $18.6$. 

Despite these results, the \textit{acc} gap with the reference setting is large on average (\eg, $40.7$\% \vs $75.1$\%), leaving many open challenges in correctly eliciting proactiveness in MLLMs. 
However, the generalization is encouraging and can serve as a starting point for future studies addressing this problem.

\section{Conclusion}
\label{sec:conclusion}
This paper presents \bench{}, a novel benchmark that evaluates MLLMs' proactiveness %
with visual inputs that require human intervention (\eg, move the occluding object) to make the query answerable.
\bench{} repurposes seven existing datasets designed for different tasks, creating sequences that evaluate proactiveness for seven distinct scenarios in single- and multi-turn interactions, both in multi-choice question answering and open-ended generation.
Our findings suggest that existing MLLMs are not proactive and prefer to abstain or hallucinate.
Additionally, our analysis shows that hinting at proactive suggestions increases proactiveness, but with marginal accuracy gains. 
Furthermore, conditioning on conversation histories and few-shot examples biases the action distribution, leading to lower accuracy.
Finally, we show that while eliciting proactiveness is challenging, learning it from data is possible. 
We release \bench{} to support future works %
on unlocking %
proactive behaviors in MLLMs.

\section*{Acknowledgements}
We acknowledge the CINECA award under the ISCRA initiative for the availability of high-performance computing resources and support. 
This work is supported by the EU projects ELIAS (No.01120237) and ELLIOT (101214398). 
Thomas De Min is funded by NextGeneration EU. 
We thank the Multimedia and Human Understanding Group (MHUG) and the Fundamental AI LAB (FunAI) for their valuable feedback and insightful suggestions.

\bibliographystyle{splncs04}
\bibliography{main}

\clearpage

\begin{center}
    \Large
    \textbf{ProactiveBench: Benchmarking Proactiveness in Multimodal Large Language Models}\\
    \vspace{5pt}
    Supplementary Material
\end{center}

\appendix
\section{Dataset details and environment implementation}
\label[appendix]{appx:dataset_details}
This section expands \cref{subsec:multichoice,subsec:dataset,subsec:filtering}, providing further information about data generation pipelines, environment details, and filtering.

\subsection{The ROD environment}
\label[appendix]{appx:subsec:rod_env}
The ROD~\cite{lee2023hardwiring} environment evaluates MLLMs' proactiveness in proposing to move occluding objects before answering the question.
The first frame in the ROD environment depicts an occluding object that completely hides another object, as \cref{fig:qualitative_rod} shows.
Each MLLM is prompted to predict the category of the occluded object, choosing out of four possible categories, and the abstain option.
As the posed question is unanswerable from the initial frame, given that the subject of the question is invisible, the environment also returns two valid proactive suggestions among other options, \ie, \texttt{move the \{occluding\_object\} to the left}, and \texttt{move the \{occluding\_object\} to the right}, where \{\texttt{occluding\_object}\} is replaced with the occluding object description (\eg, red cardboard, blue blocks).
Furthermore, we also consider camera movement a valid proactive suggestion in the free-form evaluation experiments.
A typical prompt is structured as follows:
\begin{tcolorbox}[%
    enhanced, 
    breakable,
    skin first=enhanced,
    skin middle=enhanced,
    skin last=enhanced,
    ]{}
\tt
\small
    Could you tell me what is behind the \{occluding\_object\}? \hint{} Choose from the following options. Options:\\
    A. Move the \{occluding\_object\} to the left.\\
    B. Move the \{occluding\_object\} to the right.\\
    C. \{abstain option\}.\\
    D. \{wrong random category\}.\\
    E. \{wrong random category\}.\\
    F. \{correct category\}.\\
    G. \{wrong random category\}.\\
    Please only return one of the options without any other words.
\end{tcolorbox}
The question is sampled from a pool of $15$ similar questions generated by ChatGPT, and the abstain option is from a pool of three.
Additionally, the first three options and the remaining four are shuffled, so the same option does not always appear in the same position.
Shuffling is performed during data generation, resulting in a fixed order for each sample.
Finally, \hint{} indicates the position of the hint used in the main paper experiments (\cref{subsec:analyses}), which, in the case of ROD, corresponds to ``\texttt{Hint: moving the occluding object might reveal what is behind it.}'' 

The set of valid actions $\optionst$ is constant throughout the evaluation, and MLLMs are allowed to move the occluding object $14$ times, corresponding to the total number of frames for each sample.
As the first frame is completely occluded, if a model predicts a category for the first frame, we count the prediction as wrong, as the first frame does not contain information about the target object class.
After seven consecutive right or left movements from the most occluded frame, MLLMs encounter the reference frame, where the object is perfectly visible.
Finally, the environment is circular, which means that by pursuing the same proactive suggestion, the occluding object will reveal the object until it
reappears from the opposite side, gradually re-occluding the object.

\subsection{The VSOD environment}
\label[appendix]{appx:subsec:vsod_env}
The VSOD environment evaluates MLLMs' proactiveness in proposing to wait or rewind the video before answering the question, in case of occlusions.
The first frame in this environment depicts a scene where individuals are occluded by someone passing in front of the camera, as \cref{fig:qualitative_vsod} shows.
Each MLLM is prompted to predict the speaker's name, the number of people, or the event type, choosing out of four possible categories, and the abstain option.
As the posed question is likely unanswerable from the initial frame, given that the subject of the question is (partially) invisible, the environment also returns two valid proactive suggestions among other options, \ie, \texttt{wait for the occlusion to disappear}, and \texttt{rewind the video}.
Furthermore, we also consider \textbf{camera movement} a valid proactive suggestion in the free-form evaluation experiments.
A typical prompt is structured as follows:
\begin{tcolorbox}[%
    enhanced, 
    breakable,
    skin first=enhanced,
    skin middle=enhanced,
    skin last=enhanced,
    ]{}
\tt
\small
    This is a frame extracted from a video. Answer the following question. Could you tell me who is talking? \hint{} Choose from the following options. Options:\\
    A. Rewind the video.\\
    B. \{abstain option\}.\\
    C. Wait for the occlusion to disappear.\\
    D. \{wrong random category\}.\\
    E. \{correct category\}.\\
    F. \{wrong random category\}.\\
    G. \{wrong random category\}.\\
    Please only return one of the options without any other words.
\end{tcolorbox}
In this prompt, the question is sampled from a pool of $45$ similar questions ($15$ for each question type), and the abstain option is from a pool of three.
Additionally, the first three options and the remaining four are shuffled, so the same option does not always appear in the same position.
Shuffling is performed during data generation, resulting in a fixed order for each sample.
Finally, \hint{} indicates the position of the hint used in the main paper experiments (\cref{subsec:analyses}), which, in the case of VSOD, corresponds to ``\texttt{Hint: If there is an occlusion, waiting for it to disappear or rewinding the video might reveal what's behind it}.'' 

The set of valid actions $\optionst$ is constant throughout the evaluation, and MLLMs are allowed to propose proactive suggestions as many times as the number of frames in the video.
As each occlusion lasts for a different amount of time, the number of proactive suggestions to reach a state where the question becomes answerable varies from sample to sample.
Finally, if the MLLM suggests waiting at the last frame, we treat the sequence as circular and return the first frames. 
Analogously, we return the final frame if, at the first frame, the model suggests rewinding the video.

\subsection{The MVP-N environment}
\label[appendix]{appx:subsec:mvp_n_env}
The MVP-N environment evaluates MLLMs' proactiveness in suggesting objects and camera rotations before answering the question in case of uninformative views.
The first frame in the MVP-N environment depicts an object from an uninformative viewpoint, as \cref{fig:qualitative_mvp_n} shows.
Each MLLM is prompted to predict the category of the object, choosing out of four possible categories, and the abstain option.
As the posed question is unanswerable from the initial frame, given that discriminative object features are invisible, the environment also returns a valid proactive suggestion among other options, \eg, \texttt{rotate the object}, \texttt{give me a view of the object from a different perspective}. 
As object orientation and camera extrinsic parameters are not annotated, the proactive suggestion is sampled from a pool of $11$ prompts generated with ChatGPT that contain both object rotations and camera movements.
A typical prompt is structured as follows:
\begin{tcolorbox}[%
    enhanced, 
    breakable,
    skin first=enhanced,
    skin middle=enhanced,
    skin last=enhanced,
    ]{}
\tt
\small
    Identify the object in this image. \hint{} Choose from the following options. Options:\\
    A. \{abstain option\}.\\
    B. \{proactive suggestion\}.\\
    C. \{wrong similar category\}.\\
    D. \{correct category\}.\\
    E. \{wrong similar category\}.\\
    F. \{wrong similar category\}.\\
    Please only return one of the options without any other words.
\end{tcolorbox}
In this prompt, the question is sampled from a pool of 15 similar questions, and the abstain option is from a pool of three.
Additionally, the first two options and the remaining four are shuffled, so the same option does not always appear in the same position.
Shuffling is performed during data generation, resulting in a fixed order for each sample.
Wrong option categories are sampled among those similar to the correct one to avoid leakages in the informativeness of each view.
Finally, \hint{} indicates the position of the hint used in the main paper experiments (\cref{subsec:analyses}), which, in the case of MVP-N, corresponds to ``\texttt{Hint: rotating the object could provide a more informative view.}'' 

The set of valid actions $\optionst$ is constant throughout the evaluation, and, since we generated sequences of various lengths, MLLMs are allowed to rotate the object or change camera angle $3$ times on average for each sample, depending on the sequence. 
To find the informative view, MLLMs must propose object rotations or camera movements until they reach the last state, where the object is distinguishable.

\subsection{The ImageNet-C environment}
\label[appendix]{appx:subsec:in_c_env}
The ImageNet-C environment evaluates MLLMs' proactiveness in suggesting image quality improvements before answering the question, in case of badly corrupted pictures.
The first image in the ImageNet-C environment depicts one of ImageNet~\cite{russakovsky2015imagenet} validation samples strongly corrupted by one of eight different corruptions, as \cref{fig:qualitative_in_c} shows.
Each MLLM is prompted to predict the category of the corrupted object, choosing out of four possible categories, and the abstain option.
As the posed question is hardly answerable from the initial picture, the environment also returns four proactive suggestions, out of which only one is valid, \eg, \texttt{deblur the image}, \texttt{denoise the image}, \texttt{remove artifacts}.
For example, a typical prompt is structured as follows: 
\begin{tcolorbox}[%
    enhanced, 
    breakable,
    skin first=enhanced,
    skin middle=enhanced,
    skin last=enhanced,
    ]{}
\tt
\small
    What type of object do you see here? \hint{} Choose from the following options. Options:\\
    A. \{invalid proactive suggestion\}.\\
    B. \{abstain option\}.\\
    C. \{valid proactive suggestion\}.\\
    D. \{invalid proactive suggestion\}.\\
    E. \{invalid proactive suggestion\}.\\
    F. \{wrong random category\}.\\
    G. \{correct category\}.\\
    H. \{wrong random category\}.\\
    I. \{wrong random category\}.\\
    Please only return one of the options without any other words.
\end{tcolorbox}
In this prompt, the question is sampled from a pool of $15$ similar questions, and the abstain option is from a pool of three.
Additionally, the first five options and the remaining four are shuffled, so the same option does not always appear in the same position.
Shuffling is performed during data generation, resulting in a fixed order for each sample.
Finally, \hint{} indicates the position of the hint used in the main paper experiments (\cref{subsec:analyses}), which, in the case of ImageNet-C, corresponds to ``\texttt{Hint: enhancing the image quality could help with classification.}'' 

As ImageNet-C counts $50,000$ images, we subsampled $5$ images per class, resulting in $5,000$ images, making this dataset comparable in size to the others used.
The set of valid actions $\optionst$ is constant throughout the evaluation, and MLLMs are allowed to propose the correct proactive suggestion $4$ times, improving the image quality.
After $4$ proactive suggestions, MLLMs encounter the last frame, the reference one.
Further proactive suggestions result in terminating the evaluation.

\subsection{The QuickDraw environment}
\label[appendix]{appx:subsec:qd_env}
The QuickDraw environment evaluates MLLMs' proactiveness in proposing to add details to a sketch, to make it more recognizable.
The first image in the QuickDraw environment shows the first drawn stroke by a user in trying to depict a target object, as \cref{fig:qualitative_qd} shows.
Each MLLM is prompted to predict the category of such depicted object, choosing out of four possible categories, and the abstain option.
As the posed question is likely unanswerable from the initial drawing, the environment also returns a valid proactive suggestion among other options, \eg, \texttt{add more details}, or \texttt{could you improve the quickdraw?}
For example, a typical prompt is structured as follows:
\begin{tcolorbox}[%
    enhanced, 
    breakable,
    skin first=enhanced,
    skin middle=enhanced,
    skin last=enhanced,
    ]{}
\tt
\small
    What is the category of the depicted object? \hint{} Choose from the following options. Options:\\
    A. \{proactive option\}.\\
    B. \{abstain option\}.\\
    C. \{wrong random category\}.\\
    D. \{wrong random category\}.\\
    E. \{wrong random category\}.\\
    F. \{correct category\}.\\
    Please only return one of the options without any other words.
\end{tcolorbox}
In this prompt, the question is sampled from a pool of $15$ similar questions, the abstain option is from a pool of three, and the proactive option is from a pool of $13$.
Additionally, the first two options and the remaining four are shuffled, so the same option does not always appear in the same position.
Shuffling is performed during data generation, resulting in a fixed order for each sample.
Finally, \hint{} indicates the position of the hint used in the main paper experiments (\cref{subsec:analyses}), which, in the case of QuickDraw, corresponds to ``\texttt{Hint: Adding more details to the quickdraw could help with classification.}''

As each drawing is also evaluated by a classification model~\cite{quickdraw}, we discarded all drawings not recognized by such a model, avoiding unrecognizable drawings.
Furthermore, the dataset contains $50$ million drawings over $345$ classes. 
Evaluating each MLLM would require approximately $300$ GPU days. 
Thus, we subsample it to $10$ samples per class, resulting in $3450$ drawings.
The set of valid actions $\optionst$ is constant throughout the evaluation, and MLLMs are allowed to ask for details a limited number of times, which depends on the number of strokes drawn by the user.
Depending on the number of strokes, after requesting further details enough times, MLLMs encounter the reference frame, where the object is recognizable.

\subsection{The ChangeIt environment}
\label[appendix]{appx:subsec:cit_env}
The ChangeIt environment evaluates MLLMs' proactiveness in proposing to seek the answer at a different moment in the video.
The first frame in the ChangeIt environment shows the beginning of a video tutorial, as \cref{fig:qualitative_cit} shows.
Each MLLM is prompted to either predict the category of the main object or the main action taken in the video, choosing out of four possible categories and the abstain option.
As the posed question is likely unanswerable from the initial frame, the environment also returns two valid proactive suggestions among other options, \ie, \texttt{wait for the occlusion to disappear}, and \texttt{rewind the video}.
For example, a typical prompt is structured as follows:
\begin{tcolorbox}[%
    enhanced, 
    breakable,
    skin first=enhanced,
    skin middle=enhanced,
    skin last=enhanced,
    ]{}
\tt
\small
    What action is being performed in the video? \hint{} Choose from the following options. Options:\\
    A. Rewind the video.\\
    B. Wait for the occlusion to disappear.\\
    C. \{abstain option\}.\\
    D. \{wrong random category\}.\\
    E. \{wrong random category\}.\\
    F. \{wrong random category\}.\\
    G. \{correct category\}.\\
    Please only return one of the options without any other words.
\end{tcolorbox}
For this prompt, questions related to the object category are sampled from a pool of $15$ similar questions, while those related to the action category are from a pool of $11$ questions, all obtained by querying ChatGPT.
The abstain option, instead, is sampled from a pool of three.
Additionally, the first three options and the remaining four are shuffled, so the same option does not always appear in the same position.
Shuffling is performed during data generation, resulting in a fixed order for each sample.
Finally, \hint{} indicates the position of the hint used in the main paper experiments (\cref{subsec:analyses}), which, in the case of ChangeIt, corresponds to ``\texttt{Hint: If you cannot answer the question, waiting for it to appear or rewinding the video could help with classification.}''

The set of valid actions $\optionst$ changes throughout the evaluation.
Since the environment returns the initial frame first, the rewind option is disabled at the first frame and enabled from the second step.
MLLMs can propose proactive suggestions as many times as the number of frames in the video.
Finally, as each video differs, the number of proactive suggestions to reach a state where the question becomes answerable varies from sample to sample.

\subsection{The MS-COCO environment}
\label[appendix]{appx:subsec:coco_env}
The MS-COCO environment evaluates MLLMs' proactiveness in proposing camera movements to obtain more informative cues.
The first image in the MS-COCO environment shows a trimmed picture with missing object details, as in \cref{fig:qualitative_coco}.
Since most images in MS-COCO contain multiple objects, we discard all those samples that contain more than one object, avoiding ambiguities.
Each MLLM is prompted to predict the category of the object in the image, choosing out of four possible categories and the abstain option.
For wrong option categories, we mine hard negatives using the CLIP~\cite{radford2021learning} text encoder to score the similarity between the ground truth and all the other categories.
As the posed question is likely unanswerable from the initial frame, the environment also returns one or two valid proactive suggestions, depending on how the image crop was computed.
Crops are generated to allow for exploration of one of the ordinal or cardinal directions or zooming out, the set of proactive actions, thus, changes based on the picture, \ie, \texttt{move the camera up}, \texttt{move the camera down}, \texttt{move the camera left}, \texttt{move the camera right}, and \texttt{move farther from the object}.
In the case of ordinal directions, MLLMs receive two proactive options, one for each of the cardinal directions that generate the ordinal one.
Instead, for cardinal directions and zooming out, MLLMs receive only one.
For example, a typical prompt for an ordinal direction is structured as follows:
\begin{tcolorbox}[%
    enhanced, 
    breakable,
    skin first=enhanced,
    skin middle=enhanced,
    skin last=enhanced,
    ]{}
\tt
\small
    Classify the visual content of this image. \hint{} Choose from the following options. Options:\\
    A. Move the camera left.\\
    B. Move the camera up.\\
    C. \{abstain option\}.\\
    D. \{wrong hard-negative category\}.\\
    E. \{wrong hard-negative category\}.\\
    F. \{wrong hard-negative category\}.\\
    G. \{correct category\}.\\
    Please only return one of the options without any other words.
\end{tcolorbox}
For this prompt, the question is sampled from a pool of $15$ similar questions obtained from querying ChatGPT, while the abstain option is sampled from a pool of three.
Additionally, the first two/three options (depending on the direction) and the remaining four are shuffled, so the same option does not always appear in the same position.
Shuffling is performed during data generation, resulting in a fixed order for each sample.
Finally, \hint{} indicates the position of the hint used in the main paper experiments (\cref{subsec:analyses}), which, in the case of MS-COCO, corresponds to ``\texttt{Hint: moving the camera could help with classification}'' for ordinal and cardinal directions and ``\texttt{Hint: zooming out could help with classification}'' for the zooming out case.

The set of valid actions $\optionst$ changes throughout the evaluation for ordinal directions, while it remains fixed for cardinal directions and the zooming out case.
Since the camera can move in two of the four cardinal directions in the ordinal directions case, we remove a cardinal direction if the MLLM has already unveiled all possible object details in a specific direction, \ie, it has explored all discrete steps in a direction.
Finally, MLLMs can propose proactive suggestions as many as the predefined discrete steps, set between $3$ and $5$.

\subsection{Filtering}
\label[appendix]{appx:subsec:filtering_extended}
This section provides further details about the filtering procedure, highlighting the motivations behind the proposed methodology and the effect of filtering.

As the main paper describes, filtering is necessary to avoid measuring proactiveness on samples that do not require it.
However, because filtering uses evaluated MLLMs to tell whether a sample leaks the answer in the first step, we aim to mitigate model dependency.
To do so, we employ $22$ models from $10$ families, spanning closed and open-weights and sizes ($[0.5\text{B}, >\!\!1\text{T}]$ params).
Furthermore, filtering focuses on first-turn leakages, which are easy to detect since they involve predicting the correct answer out of four candidates and cannot be used to measure proactiveness.
Thus, our filtering pipeline is not strictly dependent on the MLLM filtering pool and will not impact future MLLMs, as they will be evaluated on the same data.

We consider a sample to leak the answer if at least $25$\% ($6/22$) of MLLMs used for probing leakages can correctly answer the question in the first turn.
The $25$\% threshold is based on its high recall in identifying information-leaking samples.
Assuming a first-turn success rate of $40$\% (see \cref{tab:zero_shot_unfiltered}), this strategy achieves a $92.8$\% recall under the binomial model (\vs, $37.6$\% of the binomial test at $0.05$ confidence threshold).
Thus, this strategy mostly retains samples that require proactiveness.

\begin{figure*}[!ht]
    \centering
    \includegraphics[width=\linewidth]{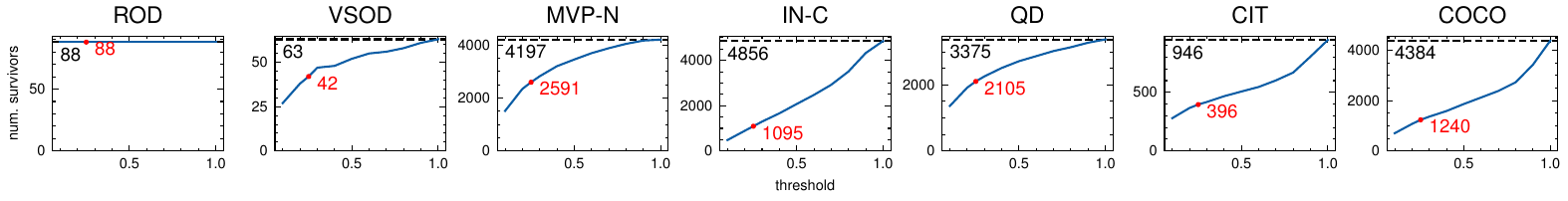}
    \caption{\textbf{Samples after filtering.} Each plot shows the remaining examples for each dataset after filtering. The light blue line represents the number of remaining examples at different thresholds. Instead, in black and red we report the original and post-filtering dataset size, respectively.}
    \label{fig:filtering_survivors}
\end{figure*}

\begin{figure*}[!ht]
    \centering
    \includegraphics[width=\linewidth]{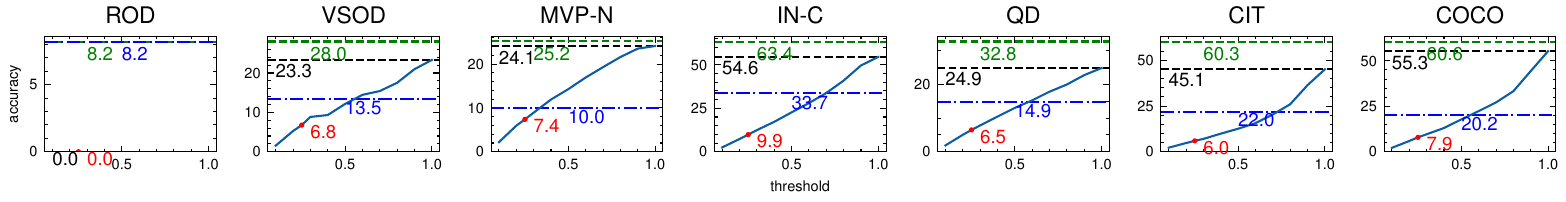}
    \caption{\textbf{Accuracy pre- and post-filtering.} Each plot shows the average MLLM's accuracy in the first turn for different datasets. The light blue line represents the accuracy at different thresholds, while in black and red we report the original and post-filtering accuracy, respectively. Finally, we report the multi-turn accuracy before and after filtering in green and blue.}
    \label{fig:filtering_accuracy}
\end{figure*}

\Cref{fig:filtering_survivors} shows for each dataset the original dataset size and the size after filtering.
Datasets with images that are generally easier to classify correctly in the first turn undergo a larger reduction (\eg, IN-C decreases from $4,856$ to $1,095$ samples).
Instead, \cref{fig:filtering_accuracy} reports, for each dataset, the average accuracy at the first turn pre- and post-filtering, and compares them with the original and post-filtering zero-shot accuracy over multiple rounds.
Finally, \cref{tab:zero_shot_unfiltered} reports MLLMs' zero-shot performance on the unfiltered benchmark.

\begin{table*}[!ht]
    \scriptsize
    \centering
    \caption{
    \textbf{MCQA results on unfiltered \bench{}.} Accuracy (\textit{acc}) and proactive suggestion rate (\textit{psr}) of MLLMs across all \bench{} splits without filtering.
    }
    \resizebox{\linewidth}{!}{
    \begin{tabular}{ 
        rlcccccccccccccccc
    }
      &  & \multicolumn{2}{c}{ROD} & \multicolumn{2}{c}{VSOD} & \multicolumn{2}{c}{MVP-N} & \multicolumn{2}{c}{IN-C} & \multicolumn{2}{c}{QD} & \multicolumn{2}{c}{CIT} & \multicolumn{2}{c}{COCO} & \multicolumn{2}{c}{avg.} \\
family & model & \textit{acc} & \textit{psr} & \textit{acc} & \textit{psr} & \textit{acc} & \textit{psr} & \textit{acc} & \textit{psr} & \textit{acc} & \textit{psr} & \textit{acc} & \textit{psr} & \textit{acc} & \textit{psr} & \textit{acc} & \textit{psr}\\
\toprule
LLaVA-1.5 & 7B & 12.5 & 0.7 & 41.3 & 1.3 & 27.7 & 0.0 & 59.4 & 0.4 & 43.0 & 0.5 & 70.3 & 0.7 & 67.6 & 0.4 & 46.0 & 0.6 \\
\rowcolor{colorline} & Mistral-7B & 0.0 & 0.0 & 9.5 & 0.2 & 13.7 & 0.1 & 53.9 & 0.2 & 12.2 & 0.1 & 46.3 & 1.4 & 49.1 & 0.0 & 26.4 & 0.3 \\
\rowcolor{colorline}\multirow{-2}{*}{LLaVA-NeXT} & Vicuna-7B & 19.3 & 0.7 & 25.4 & 0.9 & 26.2 & 0.1 & 69.2 & 0.5 & 22.0 & 0.7 & 68.6 & 0.4 & 67.7 & 0.1 & 42.6 & 0.5 \\
 & 0.5B & 44.3 & 2.3 & 20.6 & 1.9 & 30.7 & 0.4 & 53.6 & 0.7 & 45.8 & 1.1 & 59.0 & 0.6 & 61.0 & 0.1 & 45.0 & 1.0 \\
 LLaVA-OV & 7B & 0.0 & 0.0 & 30.2 & 0.3 & 24.2 & 0.0 & 70.3 & 0.4 & 46.7 & 0.3 & 56.4 & 0.1 & 60.0 & 0.0 & 41.1 & 0.2 \\
 & 72B & 0.0 & 0.0 & 41.3 & 0.3 & 23.7 & 0.0 & 74.6 & 0.4 & 39.0 & 0.1 & 61.9 & 0.2 & 59.7 & 0.0 & 42.9 & 0.1 \\
\rowcolor{colorline}SmolVLM2 & 2.2B & 0.0 & 0.0 & 23.8 & 0.3 & 26.6 & 0.0 & 55.8 & 0.5 & 27.0 & 0.5 & 64.0 & 0.3 & 59.9 & 0.0 & 36.7 & 0.2 \\
Idefics3 & 8B & 31.8 & 1.6 & 31.7 & 2.1 & 27.7 & 0.1 & 70.0 & 0.4 & 27.9 & 0.5 & 58.0 & 0.2 & 62.2 & 0.1 & 44.2 & 0.7 \\
\rowcolor{colorline}InstructBLIP & 7B & 0.0 & 0.0 & 12.7 & 1.5 & 12.8 & 0.1 & 18.9 & 0.1 & 26.0 & 0.1 & 47.6 & 0.1 & 26.7 & 0.0 & 20.7 & 0.3 \\
 & 3B & 0.0 & 0.0 & 31.7 & 0.0 & 25.4 & 0.0 & 69.5 & 0.9 & 29.1 & 0.1 & 58.9 & 0.2 & 56.6 & 0.0 & 38.7 & 0.2 \\
 & 7B & 0.0 & 0.0 & 17.5 & 0.0 & 24.7 & 0.0 & 78.5 & 0.5 & 34.3 & 0.1 & 60.6 & 0.0 & 59.5 & 0.0 & 39.3 & 0.1 \\
 & 32B & 0.0 & 0.0 & 20.6 & 0.0 & 24.9 & 0.0 & 73.6 & 0.1 & 36.4 & 0.0 & 64.1 & 0.2 & 58.1 & 0.0 & 39.7 & 0.0 \\
\multirow{-4}{*}{Qwen-2.5-VL} & 72B & 0.0 & 0.0 & 20.6 & 0.6 & 27.4 & 0.0 & 72.0 & 0.3 & 25.3 & 0.0 & 55.0 & 0.1 & 55.1 & 0.0 & 36.5 & 0.1 \\
\rowcolor{colorline} & 1B & \textbf{61.4} & 2.1 & 39.7 & 0.2 & 29.3 & 0.3 & 69.4 & 0.5 & 29.2 & 0.4 & 61.3 & 0.1 & 69.6 & 0.0 & \textbf{51.4} & 0.5 \\
\rowcolor{colorline} & 2B & 1.1 & 0.0 & \textbf{49.2} & 0.2 & 30.5 & 0.1 & 76.9 & 0.6 & 37.9 & 0.4 & 69.3 & 0.3 & 77.1 & 0.1 & 48.9 & 0.2 \\
\rowcolor{colorline}InternVL3 & 8B & 0.0 & 0.0 & 31.7 & 0.1 & 23.2 & 0.0 & 75.9 & 0.3 & 36.0 & 0.3 & 58.3 & 0.1 & 67.1 & 0.0 & 41.7 & 0.1 \\
\rowcolor{colorline} & 38B & 0.0 & 0.0 & 44.4 & 1.7 & 31.4 & 0.1 & 84.4 & 0.2 & 39.1 & 0.3 & 68.8 & 0.5 & 77.4 & 0.0 & 49.4 & 0.4 \\
\rowcolor{colorline} & 78B & 0.0 & 0.0 & 39.7 & 0.3 & 31.7 & 0.0 & 83.4 & 0.0 & 29.5 & 0.0 & 62.9 & 0.2 & 74.9 & 0.0 & 46.0 & 0.1 \\
Phi-4-Multimodal & 6B & 1.1 & 0.0 & 27.0 & 0.7 & 29.5 & 0.0 & 66.4 & 0.7 & 42.3 & 0.3 & 66.0 & 0.2 & 64.6 & 0.1 & 42.4 & 0.3 \\
\rowcolor{colorline} & o4-mini & 0.0 & 0.0 & 23.8 & 0.4 & \textbf{34.6} & 0.0 & 80.2 & 0.1 & 42.4 & 0.0 & \textbf{71.8} & 0.4 & 96.6 & 0.0 & 49.9 & 0.1 \\
\rowcolor{colorline}\multirow{-2}{*}{OpenAI} & GPT-4.1 & 0.0 & 0.0 & 9.5 & 0.1 &	24.8 & 0.1 & \textbf{90.0} & 0.3 & 35.0 & 0.1 & 62.0 & 0.3 & \textbf{96.8} & 0.0 & 45.4 & 0.1
    \end{tabular}
    }
    \label{tab:zero_shot_unfiltered}
\end{table*}

\section{Evaluating open-ended generation}
\label[appendix]{appx:free_form}
This section reports the experimental procedure followed to evaluate multimodal LLMs on \bench{} via open-ended generation, validating the multiple-choice question-answering framework used in the main paper.
Therefore, we only provide the MLLM with the image frame, the question that the model should answer, and optionally a hint to elicit proactiveness (\ie, \texttt{``If you cannot answer this question, please tell me what I should do to help you''}).

\paragraph{Evaluation protocol.}
As evaluating free-form answers is challenging, we follow previous works~\cite{liu2023visual,fu2024blink,ma2024mmlongbench,maaz2023video,song2024moviechat,nagrani2024neptune,plizzari2025omnia} and employ LLM-as-a-judge to provide a score to each answer.
As Sec.~4.1 details, we use Qwen3-8B~\cite{yang2025qwen3} and prompt it to spot proactive suggestions and category predictions.
The following system and user prompts were used to query the judge:
\begin{tcolorbox}[%
    enhanced, 
    breakable,
    skin first=enhanced,
    skin middle=enhanced,
    skin last=enhanced,
    ]{}
    \textbf{System Prompt:}\\
    \small
    \texttt{
        You are an automatic evaluation system.\textbackslash n \\\textbackslash n \\You will receive:\textbackslash n \\- A user prompt\textbackslash n \\- A list of correct answers (JSON list of strings)\textbackslash n \\- A system output\textbackslash n \\\textbackslash n \\Your task:\textbackslash n \\For each correct answer, determine whether the system output expresses the same idea, action, or requirement.\textbackslash n \\\textbackslash n \\Evaluation principles:\textbackslash n \\\textbackslash n \\1. Semantic equivalence is sufficient.\textbackslash n \\   - The wording does NOT need to match exactly.\textbackslash n \\   - Functional equivalence counts as correct.\textbackslash n \\   - If the system output describes an action that necessarily implies the correct answer, count it as present.\textbackslash n \\\textbackslash n \\2. Implicit but clear implications count as correct.\textbackslash n \\   - If the output describes the mechanism required to achieve the correct answer’s goal, count it as correct.\textbackslash n \\   - Example: "change perspective" can imply "move the camera."\textbackslash n \\\textbackslash n \\3. Do NOT require exact phrasing.\textbackslash n \\\textbackslash n \\4. Only mark 0 if:\textbackslash n \\   - The idea is clearly absent\textbackslash n \\   - The idea is contradicted\textbackslash n \\   - The answer is negated\textbackslash n \\\textbackslash n \\Procedure:\textbackslash n \\1. Think step by step.\textbackslash n \\2. For each correct answer, decide if it is semantically expressed or clearly implied.\textbackslash n \\3. Output the result on a new line in the exact format below.\textbackslash n \\\textbackslash n \\<comma-separated list of 0s and 1s>\textbackslash n \\\textbackslash n \\The last line must not contain anything else.
    }
    \tcblower
    \textbf{User Prompt:}\\
    \small
    \texttt{%
    \#\#\# User Prompt:\textbackslash n\\ \{user\_prompt\}\textbackslash n\\
    \#\#\# Correct Answers (List):\textbackslash n\\ \{correct\_answers\}\textbackslash n\\
    \#\#\# System Output:\textbackslash n\\ \{generated\_answer\}%
    }
\end{tcolorbox}
As answers are usually long, the LLM-as-a-judge is tasked to spot whether the answer contains correct proactive suggestions and correct category predictions, respectively defined in \texttt{\{correct\_answer\}}.
The judge first outputs its reasoning in \texttt{<think>} tags, then returns comma-separated values, with one digit for each correct answer (either a valid proactive suggestion or the correct category).

\section{Training details}
\label[appendix]{appx:hyperparams}
This section reports the implementation details of the RL post-training experiment described in \cref{sec:learning_proactiveness}.
The training dataset consists of approximately $27$k examples, of which $17$k are drawn from QuickDraw, with the remainder coming from MS-COCO.
We use the Huggingface GRPO trainer for our experiments, with the hyperparameters summarized in \cref{tab:hyperparams}.
Training \qwen{}-3B was performed on $4$ NVIDIA A100 GPUs, while \llavanext{}-Mistral-8B was trained on $16$ A100 GPUs due to its larger model size.
Training lasted for about $8$-$10$ hours in total.
\begin{table}
    \centering
    \caption{\textbf{RL post-training hyperparameters.}}
    \begin{tabular}{l c}
        hyperparam & value \\
    \toprule
        algorithm & GRPO \\
        batch size & $512$ \\
        optimizer & AdamW \\
        learning rate & $2\times10^{-5}$ \\
        weight decay & $0$ \\
        scheduler & cosine \\
        warmup steps & $0$ \\
        num.\ rollouts & $8$ \\
        epochs & $1$ \\
        deepspeed conf.\ & zero 3 \\
        LoRA rank & $16$ \\
        LoRA $\alpha$ & $16$ \\
        LoRA dropout & $0.1$ \\
        LoRA modules & \texttt{q\_proj}, \texttt{k\_proj} \\
    \end{tabular}
    \label{tab:hyperparams}
\end{table}

\section{Dataset examples}
\begin{figure}[!ht]
    \centering
    \includegraphics[width=\linewidth]{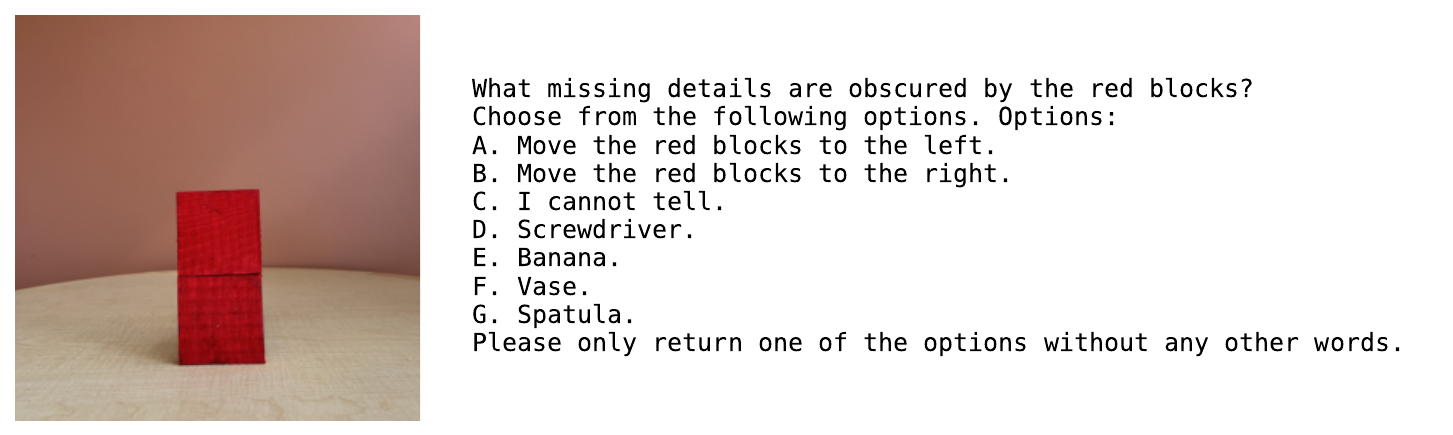}
    \caption{\textbf{ROD input example}. In the first step, the ROD environment returns images of completely occluded target objects.}
    \label{fig:qualitative_rod}
\end{figure}

\begin{figure}[!ht]
    \centering
    \includegraphics[width=\linewidth]{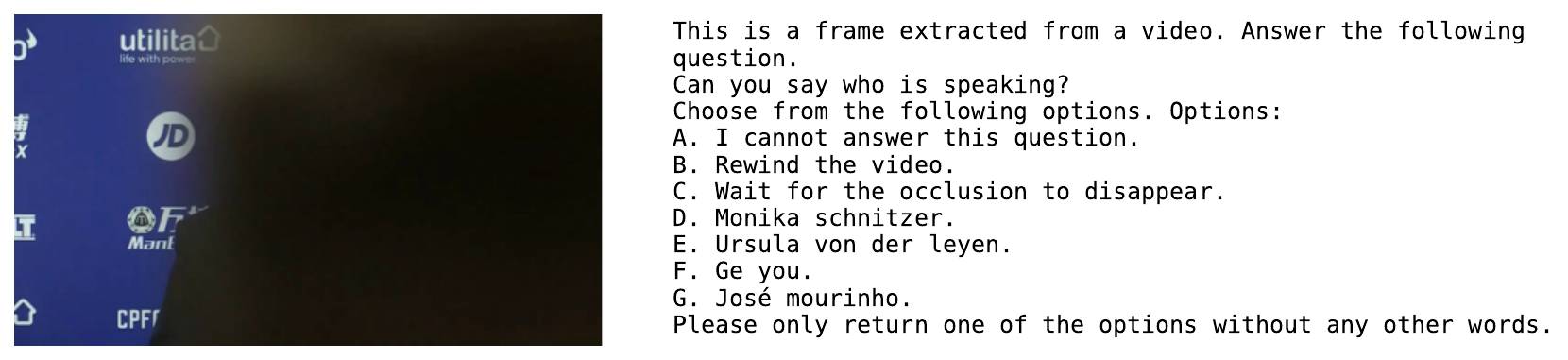}
    \caption{\textbf{VSOD input example}. In the first step, the VSOD environment returns video frames of occluded subjects.}
    \label{fig:qualitative_vsod}
\end{figure}

\begin{figure}[!ht]
    \centering
    \includegraphics[width=\linewidth]{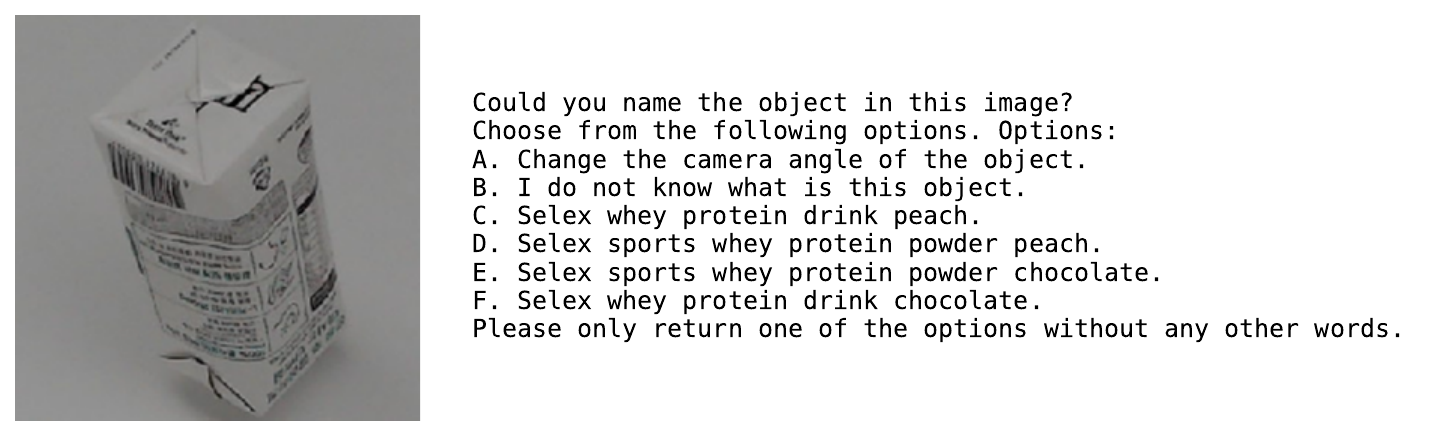}
    \caption{\textbf{MVP-N input example}. In the first step, the MVP-N environment returns uninformative object views.}
    \label{fig:qualitative_mvp_n}
\end{figure}

\begin{figure}[!ht]
    \centering
    \includegraphics[width=\linewidth]{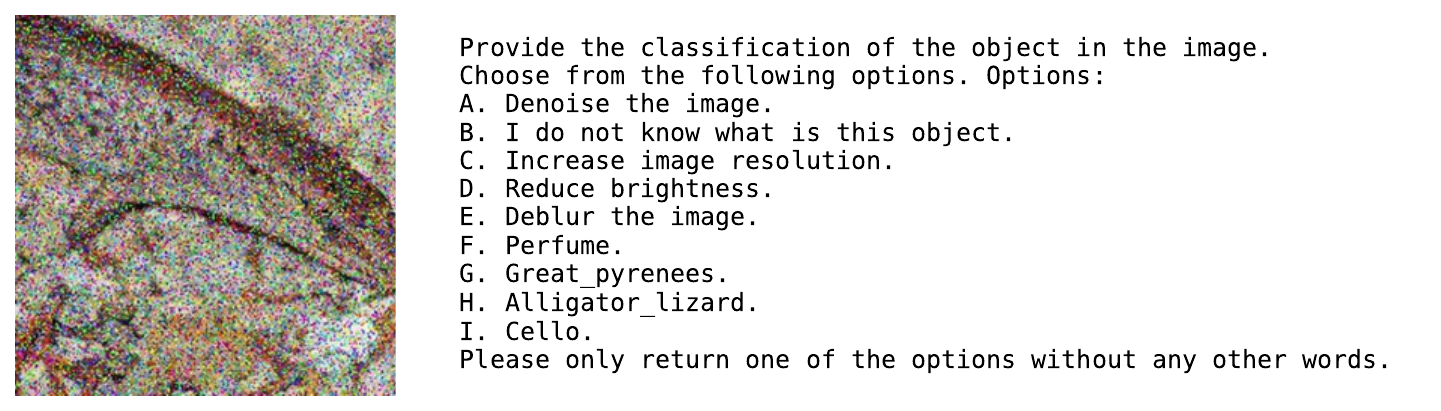}
    \caption{\textbf{ImageNet-C input example}. In the first step, the IN-C environment returns heavily corrupted images.}
    \label{fig:qualitative_in_c}
\end{figure}

\begin{figure}[!ht]
    \centering
    \includegraphics[width=\linewidth]{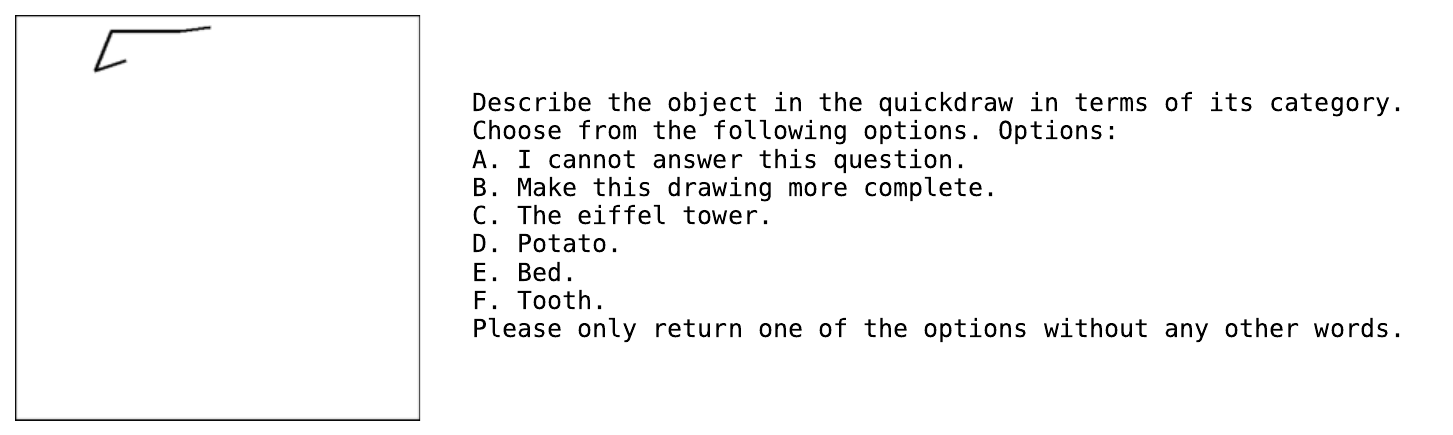}
    \caption{\textbf{QuickDraw input example}. In the first step, the QD environment returns the first stroke of a sketch.}
    \label{fig:qualitative_qd}
\end{figure}

\begin{figure}[!ht]
    \centering
    \includegraphics[width=\linewidth]{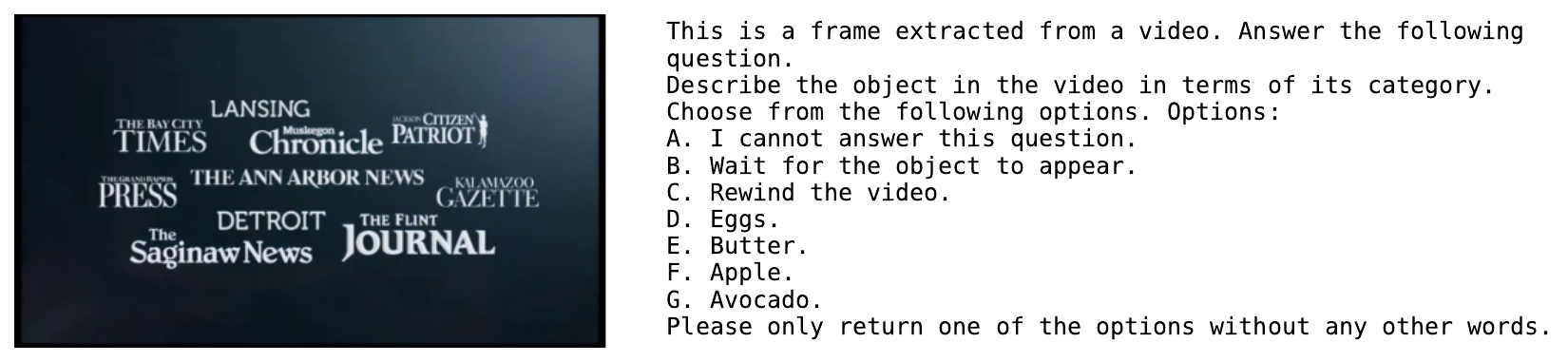}
    \caption{\textbf{ChangeIt input example}. In the first step, the CIT environment returns video frames where the target object or action will appear in the future.}
    \label{fig:qualitative_cit}
\end{figure}

\begin{figure}[!ht]
    \centering
    \includegraphics[width=\linewidth]{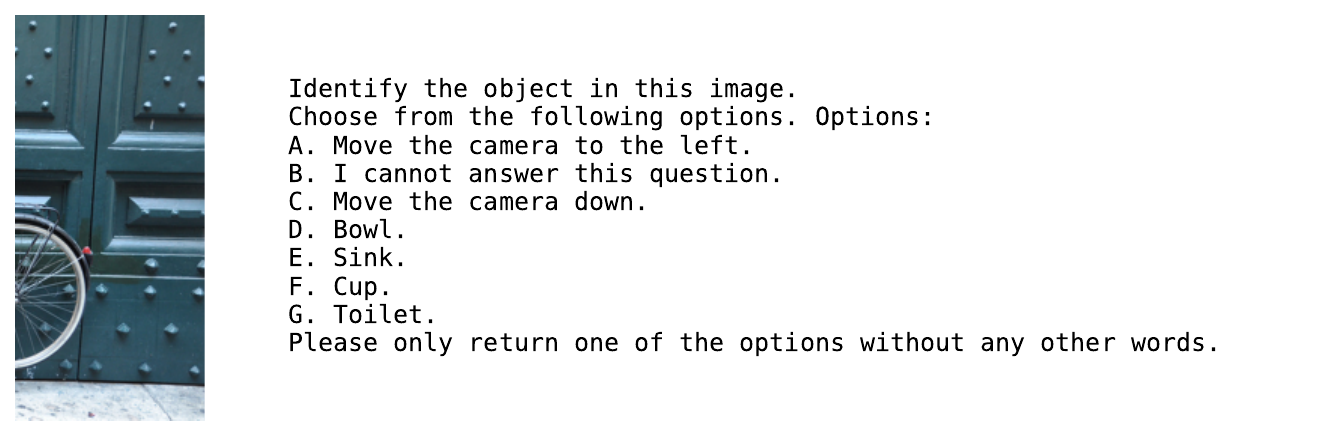}
    \caption{\textbf{MS-COCO input example}. In the first step, the COCO environment returns images where object details are removed.}
    \label{fig:qualitative_coco}
\end{figure}

\label[appendix]{appx:qualitatives}
\Cref{fig:qualitative_rod,fig:qualitative_vsod,fig:qualitative_mvp_n,fig:qualitative_in_c,fig:qualitative_qd,fig:qualitative_cit,fig:qualitative_coco} report dataset examples returned by the environment in the first state.


\section{Extended results}
\begin{figure*}[!ht]
    \centering
    \includegraphics[width=\linewidth]{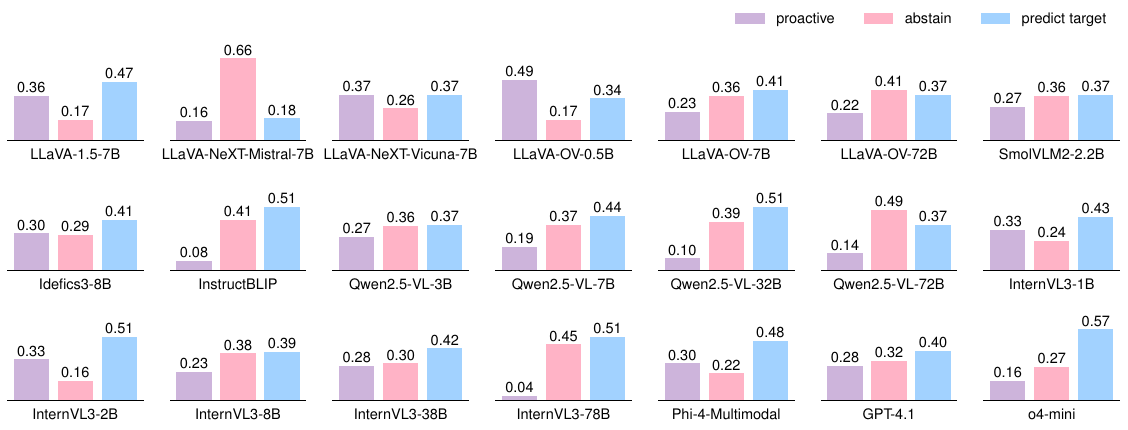}
    \caption{\textbf{Action distributions.} We report the action distribution for all evaluated models.}
    \label{fig:dist_zero_shot_full}
\end{figure*}

\begin{figure*}[!ht]
    \centering
    \includegraphics[width=\linewidth]{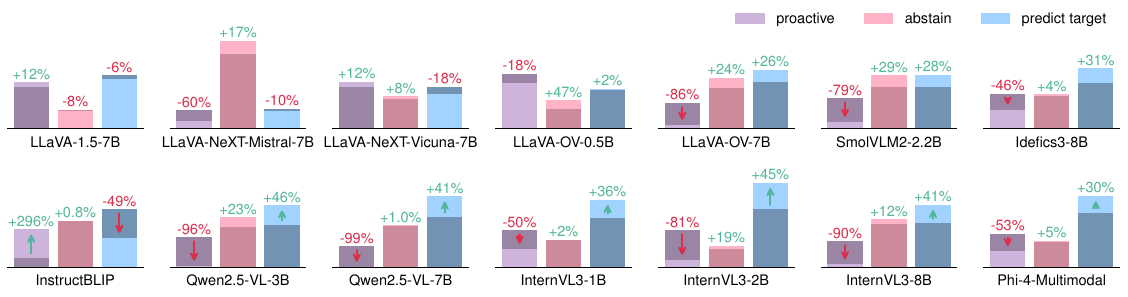}
    \caption{\textbf{Action distributions with random proactive options.} Lighter bars describe variations using random proactive suggestions for all evaluated models.}
    \label{fig:dist_random_full}
\end{figure*}

\begin{figure*}[!ht]
    \centering
    \includegraphics[width=\linewidth]{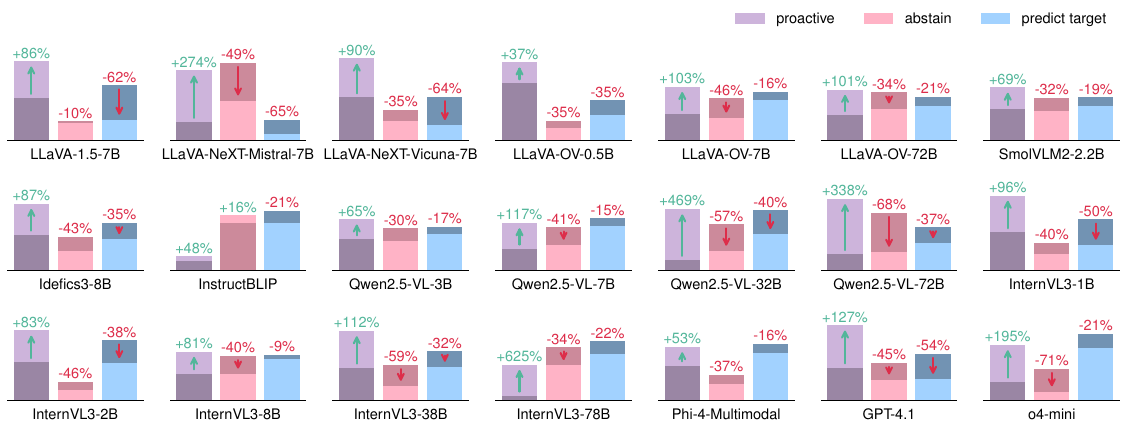}
    \caption{\textbf{Action distributions with hints.} Bars describe action distributions with (light) and without (dark) hints in the prompt for all evaluated models.}
    \label{fig:dist_hint_full}
\end{figure*}

\label[appendix]{appx:extended_results}
As most results could not fit within nine pages, the main paper summarizes key findings with plots.
This section reports all tables associated with the main paper's plots and the extended version of each plot, not limited to six models.
\Cref{tab:reference} reports MLLMs oracle performance on \bench{}.
\Cref{fig:dist_zero_shot_full} shows the action distribution for all models, further highlighting that some overweight proactive suggestions over the abstain option.
\internvl{} 78B stands out, showing the lowest rate of proactive suggestions ($4$\%), despite being one of the best open-weight MLLMs.
\Cref{fig:dist_random_full} reports MLLM's action distribution when proactive suggestions are replaced with random ones.
Similarly, \cref{tab:hint} and \cref{fig:dist_hint_full} describe MLLM's results and action distribution on all models when the prompt hints at proactive suggestions.
\Cref{tab:free_form_full,tab:free_form_hint} integrate the MLLMs' open-ended generation results, with and without hint conditioning, with the \textit{agg} metric, which aggregates correct answers and valid proactive suggestions into a single score.
Finally, \cref{tab:free_form_multi_turn} shows MLLMs' multi-turn OEG results by mapping inapplicable suggestions to valid ones.

\paragraph{Computational details.}
We conducted most experiments using a single A100 Nvidia GPU, $32$GB of RAM, and $8$ CPU cores, lasting about $1$ hour, depending on the dataset.
When conditioning on conversation histories and few-shot samples, we used two A100 GPUs to reduce the memory footprint of the models' parameters, with experiments lasting about $2$ hours on average and at most $8$ hours, depending on the dataset and model.
Furthermore, to avoid out-of-memory issues for \phimultimodal{} with ICL examples, we reduced the ROD image sizes of the few shots from $3024\times3024$ to $512\times512$, and the sequence length of MVP-N to $2$ when using $3$ shots.
Finally, we resized all samples' short edge to $224$px when conditioning on conversational histories to avoid out-of-memory issues with long sequences.

\begin{table*}[!ht]
    \scriptsize
    \centering
    \caption{\textbf{MCQA oracle results on \bench{}.} Accuracy of MLLMs across all \bench{} splits.}
    \begin{tabularx}{\linewidth} { 
        R{2.3cm} %
        L{1.4cm} %
        *{8}{Y}
    }
    family & model & \multicolumn{1}{c}{ROD} & \multicolumn{1}{c}{VSOD} & \multicolumn{1}{c}{MVP-N} & \multicolumn{1}{c}{IN-C} & \multicolumn{1}{c}{QD} & \multicolumn{1}{c}{CIT} & \multicolumn{1}{c}{COCO} & \multicolumn{1}{c}{avg.} \\
\toprule
LLaVA-1.5 & 7B & \textbf{100.0} & 76.2 & 32.6 & 91.0 & 72.9 & 76.8 & 93.0 & 77.5 \\
\rowcolor{colorline} & Mistral-7B & \textbf{100.0} & 57.1 & 43.6 & 88.6 & 65.6 & 75.3 & 95.5 & 75.1 \\
\rowcolor{colorline}\multirow{-2}{*}{LLaVA-NeXT} & Vicuna-7B & 98.9 & 57.1 & 36.6 & 90.6 & 56.8 & 74.2 & 95.2 & 72.8 \\
 & 0.5B & \textbf{100.0} & 40.5 & 60.7 & 84.6 & 78.1 & 78.5 & 96.0 & 76.9 \\
 LLaVA-OV & 7B & \textbf{100.0} & 78.6 & 63.2 & 94.5 & 86.4 & 87.4 & 97.6 & 86.8 \\
 & 72B & \textbf{100.0} & 83.3 & 68.0 & 95.3 & 88.1 & 87.1 & 97.6 & 88.5 \\
\rowcolor{colorline}SmolVLM2 & 2.2B & \textbf{100.0} & 69.0 & 50.4 & 88.9 & 73.1 & 84.6 & 95.8 & 80.3 \\
Idefics3 & 8B & \textbf{100.0} & 76.2 & 52.5 & 90.4 & 67.5 & 83.6 & 96.1 & 80.9 \\
\rowcolor{colorline}InstructBLIP & 7B & 75.0 & 57.1 & 21.5 & 31.5 & 32.5 & 61.4 & 25.6 & 43.5 \\
 & 3B & \textbf{100.0} & 78.6 & 51.7 & 91.5 & 65.5 & 84.8 & 96.0 & 81.2 \\
 & 7B & \textbf{100.0} & 81.0 & 63.3 & 95.0 & 75.3 & \textbf{87.9} & 97.1 & 85.6 \\
 & 32B & \textbf{100.0} & 78.6 & 53.8 & 93.2 & 72.3 & 84.3 & 95.8 & 82.6 \\
\multirow{-4}{*}{Qwen-2.5-VL} & 72B & \textbf{100.0} & 76.2 & 63.8 & 94.7 & 80.3 & 84.8 & 97.3 & 85.3 \\
\rowcolor{colorline} & 1B & 98.9 & 52.4 & 55.4 & 88.0 & 62.5 & 80.3 & 96.5 & 76.3 \\
\rowcolor{colorline} & 2B & \textbf{100.0} & 76.2 & 57.0 & 92.7 & 65.8 & 84.1 & 97.6 & 81.9 \\
\rowcolor{colorline}InternVL3 & 8B & \textbf{100.0} & 76.2 & 59.9 & 96.1 & 70.1 & 84.1 & 97.6 & 83.4 \\
\rowcolor{colorline} & 38B & \textbf{100.0} & 81.0 & 72.2 & 97.5 & 80.7 & 86.1 & 97.8 & 87.9 \\
\rowcolor{colorline} & 78B & \textbf{100.0} & \textbf{85.7} & 74.5 & \textbf{98.3} & 80.9 & 87.4 & \textbf{98.7} & \textbf{89.4} \\
Phi-4-Multimodal & 6B & \textbf{100.0} & 57.1 & 47.5 & 82.6 & 77.9 & 74.2 & 96.0 & 76.5 \\
\rowcolor{colorline} & GPT-4.1 & \textbf{100.0} & 76.2 & \textbf{80.8} & 98.2 & \textbf{88.2} & 83.6 & 96.8 & 89.1 \\
\rowcolor{colorline}\multirow{-2}{*}{OpenAI} & o4-mini & 92.0 & 64.3 & 73.4 & 65.4 & 64.4 & 65.4 & 92.6 & 73.9 \\
    \end{tabularx}
    \label{tab:reference}
\end{table*}

\begin{table*}[!ht]
    \scriptsize
    \centering
    \caption{\textbf{MCQA results on \bench{} by hinting at proactive suggestions.} Accuracy (\textit{acc}) and proactive suggestion rate (\textit{psr}) of MLLMs across all \bench{} splits.
    }
    \resizebox{\linewidth}{!}{
    \begin{tabular}{ 
        rlcccccccccccccccc
    }
     &  & \multicolumn{2}{c}{ROD} & \multicolumn{2}{c}{VSOD} & \multicolumn{2}{c}{MVP-N} & \multicolumn{2}{c}{IN-C} & \multicolumn{2}{c}{QD} & \multicolumn{2}{c}{CIT} & \multicolumn{2}{c}{COCO} & \multicolumn{2}{c}{avg.} \\
family & model & \textit{acc} & \textit{psr} & \textit{acc} & \textit{psr} & \textit{acc} & \textit{psr} & \textit{acc} & \textit{psr} & \textit{acc} & \textit{psr} & \textit{acc} & \textit{psr} & \textit{acc} & \textit{psr} & \textit{acc} & \textit{psr}\\
\toprule
LLaVA-1.5 & 7B & 47.7 & 4.9 & 28.6 & 25.9 & 14.7 & 2.3 & 11.1 & 1.3 & 37.9 & 1.8 & 48.7 & 2.2 & 41.0 & 1.5 & 32.8 & 5.7 \\
\rowcolor{colorline} & Mistral-7B & 2.3 & 3.7 & 0.0 & 4.7 & 2.5 & 2.4 & 14.3 & 1.1 & 4.8 & 1.0 & 5.8 & 1.1 & 11.3 & 0.5 & 5.9 & 2.1 \\
\rowcolor{colorline}\multirow{-2}{*}{LLaVA-NeXT} & Vicuna-7B & 44.3 & 4.9 & 14.3 & 43.3 & 14.0 & 2.3 & 17.8 & 1.2 & 10.4 & 2.3 & 50.8 & 3.4 & 46.4 & 1.3 & 28.3 & 8.4 \\
 & 0.5B & 44.3 & 5.6 & 9.5 & 29.4 & 17.7 & 1.0 & 19.1 & 1.9 & 36.5 & 2.1 & 44.2 & 6.5 & 38.9 & 1.1 & 30.0 & 6.8 \\
 LLaVA-OV & 7B & 20.5 & 0.6 & 23.8 & 0.7 & 27.7 & 1.0 & 40.6 & 2.1 & 28.5 & 0.5 & 8.8 & 0.5 & 9.2 & 0.2 & 22.7 & 0.8 \\
 & 72B & 0.0 & 0.0 & 14.3 & 0.1 & 19.6 & 0.7 & 41.2 & 2.0 & 20.1 & 0.6 & 14.4 & 0.5 & 14.4 & 0.3 & 17.7 & 0.6 \\
\rowcolor{colorline}SmolVLM2 & 2.2B & 0.0 & 0.1 & 14.3 & 0.2 & 16.1 & 0.5 & 29.2 & 2.2 & 11.2 & 0.8 & 51.3 & 1.8 & 7.8 & 0.1 & 18.6 & 0.8 \\
Idefics3 & 8B & 29.5 & 9.4 & 28.6 & 37.3 & 13.7 & 0.7 & 33.2 & 0.9 & 15.9 & 1.4 & 24.7 & 0.9 & 34.4 & 1.0 & 25.7 & 7.4 \\
\rowcolor{colorline}InstructBLIP & 7B & 1.1 & 0.5 & 16.7 & 4.9 & 7.4 & 0.1 & 7.9 & 0.1 & 14.2 & 0.1 & 22.7 & 0.2 & 10.0 & 0.0 & 11.4 & 0.8 \\
 & 3B & 48.9 & 1.3 & 33.3 & 4.2 & 12.6 & 0.5 & 33.8 & 2.6 & 11.1 & 0.5 & 11.9 & 0.6 & 10.5 & 0.1 & 23.1 & 1.4 \\
 & 7B & 0.0 & 0.0 & 9.5 & 0.1 & 12.6 & 0.3 & 50.0 & 2.1 & 23.8 & 0.9 & 6.3 & 0.2 & 6.3 & 0.0 & 15.5 & 0.5 \\
 & 32B & 10.2 & 0.8 & 2.4 & 0.2 & 25.4 & 1.2 & 40.9 & 1.4 & 26.4 & 1.1 & 15.7 & 0.9 & 24.1 & 0.6 & 20.7 & 0.9 \\
\multirow{-4}{*}{Qwen-2.5-VL} & 72B & 0.0 & 0.5 & 9.5 & 0.6 & 28.2 & 1.3 & 44.7 & 2.4 & 26.8 & 1.9 & 23.7 & 1.1 & 32.1 & 0.9 & 23.6 & 1.2 \\
\rowcolor{colorline} & 1B & \textbf{62.5} & 2.9 & 23.8 & 1.4 & 33.0 & 1.6 & 26.4 & 2.0 & 25.7 & 2.3 & 39.9 & 2.1 & 31.0 & 0.7 & 34.6 & 1.9 \\
\rowcolor{colorline} & 2B & 42.0 & 1.1 & 52.4 & 14.4 & 34.4 & 1.1 & 41.3 & 1.5 & 29.2 & 1.8 & 32.8 & 2.5 & 52.8 & 0.9 & 40.7 & 3.3 \\
\rowcolor{colorline}InternVL3 & 8B & 2.3 & 0.0 & 19.0 & 0.1 & 19.0 & 0.6 & 38.6 & 1.1 & 26.7 & 1.1 & 9.6 & 0.3 & 13.7 & 0.2 & 18.4 & 0.5 \\
\rowcolor{colorline} & 38B & 3.4 & 0.0 & 33.3 & 2.0 & 38.2 & 1.3 & 53.5 & 1.9 & 35.7 & 1.8 & 29.8 & 1.5 & 56.0 & 0.9 & 35.7 & 1.3 \\
\rowcolor{colorline} & 78B & 0.0 & 0.0 & 31.0 & 1.4 & 18.2 & 0.3 & 57.0 & 0.8 & 11.4 & 0.4 & 29.0 & 1.3 & 23.9 & 0.2 & 24.3 & 0.6 \\
Phi-4-Multimodal & 6B & 9.1 & 0.4 & 9.5 & 1.1 & 21.4 & 0.2 & 35.1 & 2.3 & 34.3 & 1.3 & 23.2 & 0.5 & 26.0 & 0.4 & 22.7 & 0.9 \\
\rowcolor{colorline} & GPT-4.1 & 0.0 & 0.0 & 7.1 & 0.8 & \textbf{52.2} & 2.8 & 30.8 & 3.0 & 33.0 & 2.5 & 24.7 & 0.7 & 92.8 & 0.1 & 34.4 & 1.4 \\
\rowcolor{colorline} & GPT-5.2 & 1.1 & 0.0 & 0.0 & 0.8 & 20.8 & 1.6 & 35.8 & 1.0 & 41.4 & 1.6 & 35.4 & 1.1 & 94.4 & 0.0 & 32.7 & 0.9 \\
\rowcolor{colorline}\multirow{-2}{*}{OpenAI} & o4-mini & 20.5 & 0.2 & \textbf{54.8} & 4.6 & 31.4 & 0.5 & \textbf{66.8} & 0.8 & \textbf{62.0} & 1.3 & \textbf{59.8} & 1.8 & \textbf{94.6} & 0.1 & \textbf{55.7} & 1.3 \\
    \end{tabular}
    }
    \label{tab:hint}
\end{table*}

\begin{table*}[t]
    \centering
    \scriptsize
    \caption{\textbf{OEG results on \bench{}.} We report the aggregate accuracy (\textit{agg}), the ratio of correctly predicted categories (\textit{cc}), and the ratio of correct proactive suggestions (\textit{ps}) for all datasets, with global averages in the last column.}
    \begin{tabularx}{\linewidth} { 
        R{1.8cm} %
        L{1.4cm} %
        *{12}{Y}
    }
    &  & \multicolumn{3}{c}{ROD} & \multicolumn{3}{c}{VSOD} & \multicolumn{3}{c}{MVP-N} & \multicolumn{3}{c}{IN-C} \\
 family & model & \textit{agg} & \textit{cc} & \textit{ps} & \textit{agg} & \textit{cc} & \textit{ps} & \textit{agg} & \textit{cc} & \textit{ps} & \textit{agg} & \textit{cc} & \textit{ps} \\
\toprule
 LLaVA-1.5 & 7B & 5.7 & 0.0 & 5.7 & 0.0 & 0.0 & 0.0 & 0.0 & 0.0 & 0.0 & 3.0 & 1.0 & 2.0 \\
 \rowcolor{colorline} & Mistral-7B & 3.4 & 0.0 & 3.4 & 2.4 & 0.0 & \textbf{2.4} & 0.0 & 0.0 & 0.0 & 34.0 & 1.0 & 34.0 \\
 \rowcolor{colorline}\multirow{-2}{*}{LLaVA-NeXT} & Vicuna-7B & 5.7 & 0.0 & 5.7 & 2.4 & 2.4 & 0.0 & 0.0 & 0.0 & 0.0 & \textbf{37.0} & 3.0 & \textbf{36.0} \\
  & 0.5B & 1.1 & 0.0 & 1.1 & 0.0 & 0.0 & 0.0 & 0.0 & 0.0 & 0.0 & 11.0 & 0.0 & 11.0 \\
  LLaVA-OV & 7B & 1.1 & 0.0 & 1.1 & 2.4 & 0.0 & \textbf{2.4} & 0.0 & 0.0 & 0.0 & 19.0 & 2.0 & 18.0 \\
  & 72B & 5.7 & 0.0 & 5.7 & 0.0 & 0.0 & 0.0 & 0.0 & 0.0 & 0.0 & 20.0 & 2.0 & 19.0 \\
 \rowcolor{colorline}SmolVLM2 & 2.2B & 0.0 & 0.0 & 0.0 & 0.0 & 0.0 & 0.0 & 0.0 & 0.0 & 0.0 & 5.0 & 0.0 & 5.0 \\
 Idefics3 & 8B & 1.1 & 0.0 & 1.1 & 2.4 & 2.4 & 0.0 & 0.0 & 0.0 & 0.0 & 7.0 & 2.0 & 7.0 \\
 \rowcolor{colorline} & 3B & 1.1 & 0.0 & 1.1 & 2.4 & 2.4 & 0.0 & 0.0 & 0.0 & 0.0 & 23.0 & 2.0 & 22.0 \\
 \rowcolor{colorline} & 7B & \textbf{6.8} & 0.0 & \textbf{6.8} & 2.4 & 2.4 & 0.0 & \textbf{1.0} & 0.0 & \textbf{1.0} & 32.0 & 3.0 & 31.0 \\
 \rowcolor{colorline} & 32B & 3.4 & 0.0 & 3.4 & 0.0 & 0.0 & 0.0 & 0.0 & 0.0 & 0.0 & 27.0 & 2.0 & 25.0 \\
 \rowcolor{colorline}\multirow{-4}{*}{Qwen-2.5-VL} & 72B & 2.3 & 0.0 & 2.3 & 0.0 & 0.0 & 0.0 & 0.0 & 0.0 & 0.0 & 29.0 & 3.0 & 28.0 \\
  & 1B & 1.1 & 0.0 & 1.1 & 2.4 & 2.4 & 0.0 & 0.0 & 0.0 & 0.0 & 19.0 & 3.0 & 17.0 \\
  & 2B & 0.0 & 0.0 & 0.0 & 2.4 & 2.4 & 0.0 & 0.0 & 0.0 & 0.0 & 19.0 & 2.0 & 17.0 \\
 InternVL3 & 8B & 1.1 & 0.0 & 1.1 & \textbf{4.8} & \textbf{4.8} & 0.0 & 0.0 & 0.0 & 0.0 & 20.0 & 3.0 & 18.0 \\
  & 38B & 3.4 & 0.0 & 3.4 & \textbf{4.8} & \textbf{4.8} & 0.0 & 0.0 & 0.0 & 0.0 & 20.0 & 8.0 & 15.0 \\
  & 78B & 1.1 & 0.0 & 1.1 & 2.4 & 2.4 & 0.0 & 0.0 & 0.0 & 0.0 & 25.0 & \textbf{11.0} & 16.0 \\
[-2ex]\\
 &  & \multicolumn{3}{c}{QD} & \multicolumn{3}{c}{CIT} & \multicolumn{3}{c}{COCO} & \multicolumn{3}{c}{avg.} \\
 family & model & \textit{agg} & \textit{cc} & \textit{ps} & \textit{agg} & \textit{cc} & \textit{ps} & \textit{agg} & \textit{cc} & \textit{ps} & \textit{agg} & \textit{cc} & \textit{ps} \\
\toprule
 LLaVA-1.5 & 7B & 7.0 & 1.0 & 6.0 & 3.0 & \textbf{3.0} & 0.0 & 1.0 & 1.0 & 1.0 & 2.8 & 0.9 & 2.1 \\
 \rowcolor{colorline} & Mistral-7B & \textbf{29.0} & \textbf{4.0} & \textbf{27.0} & 8.0 & 2.0 & 6.0 & 5.0 & 4.0 & 1.0 & \textbf{11.7} & 1.6 & 10.5 \\
 \rowcolor{colorline}\multirow{-2}{*}{LLaVA-NeXT} & Vicuna-7B & 23.0 & 0.0 & 23.0 & 9.0 & \textbf{3.0} & 6.0 & 4.0 & 0.0 & \textbf{4.0} & 11.6 & 1.2 & \textbf{10.7} \\
  & 0.5B & 4.0 & 2.0 & 2.0 & 1.0 & 1.0 & 0.0 & 1.0 & 0.0 & 1.0 & 2.6 & 0.4 & 2.2 \\
  LLaVA-OV & 7B & 9.0 & 0.0 & 9.0 & 2.0 & 0.0 & 2.0 & 4.0 & 2.0 & 2.0 & 5.4 & 0.6 & 4.9 \\
  & 72B & 8.0 & 2.0 & 7.0 & 1.0 & 0.0 & 1.0 & 1.0 & 0.0 & 1.0 & 5.1 & 0.6 & 4.8 \\
 \rowcolor{colorline}SmolVLM2 & 2.2B & 3.0 & 3.0 & 0.0 & 0.0 & 0.0 & 0.0 & 1.0 & 0.0 & 1.0 & 1.3 & 0.4 & 0.9 \\
 Idefics3 & 8B & 1.0 & 0.0 & 1.0 & 0.0 & 0.0 & 0.0 & 4.0 & 4.0 & 1.0 & 2.2 & 1.2 & 1.4 \\
 \rowcolor{colorline} & 3B & 5.0 & 2.0 & 3.0 & 1.0 & 0.0 & 1.0 & 3.0 & 3.0 & 0.0 & 5.1 & 1.3 & 3.9 \\
 \rowcolor{colorline} & 7B & 16.0 & 0.0 & 16.0 & 8.0 & 1.0 & 8.0 & 5.0 & 2.0 & 3.0 & 10.2 & 1.2 & 9.4 \\
 \rowcolor{colorline} & 32B & 3.0 & 0.0 & 3.0 & \textbf{11.0} & 2.0 & \textbf{9.0} & 0.0 & 0.0 & 0.0 & 6.3 & 0.6 & 5.8 \\
 \rowcolor{colorline}\multirow{-4}{*}{Qwen-2.5-VL} & 72B & 12.0 & 1.0 & 11.0 & 7.0 & 1.0 & 6.0 & 6.0 & 4.0 & 2.0 & 8.0 & 1.3 & 7.0 \\
  & 1B & 7.0 & 2.0 & 5.0 & 2.0 & 1.0 & 1.0 & 6.0 & 3.0 & 3.0 & 5.4 & 1.6 & 3.9 \\
  & 2B & 3.0 & 0.0 & 3.0 & 2.0 & 1.0 & 1.0 & 0.0 & 0.0 & 0.0 & 3.8 & 0.8 & 3.0 \\
 InternVL3 & 8B & 1.0 & 0.0 & 1.0 & 3.0 & 1.0 & 2.0 & 2.0 & 2.0 & 0.0 & 4.6 & 1.5 & 3.2 \\
  & 38B & 2.0 & 1.0 & 1.0 & 7.0 & 1.0 & 6.0 & 4.0 & 4.0 & 0.0 & 5.9 & 2.7 & 3.6 \\
  & 78B & 6.0 & 1.0 & 5.0 & 6.0 & \textbf{3.0} & 3.0 & \textbf{7.0} & \textbf{7.0} & 0.0 & 6.8 & \textbf{3.5} & 3.6 \\
    \end{tabularx}
    \label{tab:free_form_full}
\end{table*}

\begin{table*}[t]
    \centering
    \scriptsize
    \caption{\textbf{OEG results on \bench{} by hinting at proactive suggestions.} We report the aggregate accuracy (\textit{agg}), the ratio of correctly predicted categories (\textit{cc}), and the ratio of correct proactive suggestions (\textit{ps}) for all datasets, with global averages in the last column.}
    \begin{tabularx}{\linewidth}{ 
        R{1.8cm} %
        L{1.4cm} %
        *{16}{Y}
    }
     &  & \multicolumn{3}{c}{ROD} & \multicolumn{3}{c}{VSOD} & \multicolumn{3}{c}{MVP-N} & \multicolumn{3}{c}{IN-C} \\
 family & model & \textit{agg} & \textit{cc} & \textit{ps} & \textit{agg} & \textit{cc} & \textit{ps} & \textit{agg} & \textit{cc} & \textit{ps} & \textit{agg} & \textit{cc} & \textit{ps} \\
\toprule
 LLaVA-1.5 & 7B & 11.4 & 0.0 & 11.4 & 2.4 & \textbf{2.4} & 2.4 & 0.0 & 0.0 & 0.0 & 11.0 & 3.0 & 9.0 \\
 \rowcolor{colorline} & Mistral-7B & 17.0 & 0.0 & 17.0 & 0.0 & 0.0 & 0.0 & 3.0 & 0.0 & 3.0 & 61.0 & 0.0 & 61.0 \\
 \rowcolor{colorline}\multirow{-2}{*}{LLaVA-NeXT} & Vicuna-7B & 29.5 & 0.0 & 29.5 & 0.0 & 0.0 & 0.0 & 2.0 & 0.0 & 2.0 & 55.0 & 0.0 & 55.0 \\
  & 0.5B & 14.8 & 0.0 & 14.8 & 4.8 & \textbf{2.4} & 2.4 & 1.0 & 0.0 & 1.0 & 6.0 & 0.0 & 6.0 \\
  LLaVA-OV & 7B & 5.7 & 0.0 & 5.7 & 0.0 & 0.0 & 0.0 & 2.0 & 0.0 & 2.0 & 52.0 & 1.0 & 52.0 \\
  & 72B & 29.5 & 0.0 & 29.5 & 4.8 & 0.0 & 4.8 & 3.0 & 0.0 & 3.0 & 61.0 & 1.0 & 60.0 \\
 \rowcolor{colorline}SmolVLM2 & 2.2B & 5.7 & 0.0 & 5.7 & 0.0 & 0.0 & 0.0 & 0.0 & 0.0 & 0.0 & 5.0 & 0.0 & 5.0 \\
 Idefics3 & 8B & 3.4 & 0.0 & 3.4 & 0.0 & 0.0 & 0.0 & 0.0 & 0.0 & 0.0 & 13.0 & 1.0 & 13.0 \\
 \rowcolor{colorline} & 3B & 6.8 & 0.0 & 6.8 & 0.0 & 0.0 & 0.0 & 0.0 & 0.0 & 0.0 & 28.0 & 1.0 & 27.0 \\
 \rowcolor{colorline} & 7B & 5.7 & 0.0 & 5.7 & 0.0 & 0.0 & 0.0 & 1.0 & 0.0 & 1.0 & 72.0 & 1.0 & 72.0 \\
 \rowcolor{colorline} & 32B & 37.5 & 0.0 & 37.5 & 2.4 & 0.0 & 2.4 & 4.0 & 0.0 & 4.0 & 70.0 & 0.0 & 70.0 \\
 \rowcolor{colorline}\multirow{-4}{*}{Qwen-2.5-VL} & 72B & \textbf{63.6} & 0.0 & \textbf{63.6} & \textbf{9.5} & 0.0 & \textbf{9.5} & 1.0 & 0.0 & 1.0 & \textbf{75.0} & 0.0 & \textbf{75.0} \\
  & 1B & 6.8 & 0.0 & 6.8 & 4.8 & \textbf{2.4} & 2.4 & 1.0 & 0.0 & 1.0 & 25.0 & 0.0 & 25.0 \\
  & 2B & 10.2 & 0.0 & 10.2 & 0.0 & 0.0 & 0.0 & 0.0 & 0.0 & 0.0 & 54.0 & 2.0 & 54.0 \\
 InternVL3 & 8B & 29.5 & 0.0 & 29.5 & 2.4 & 0.0 & 2.4 & 0.0 & 0.0 & 0.0 & 52.0 & 0.0 & 52.0 \\
  & 38B & 25.0 & 0.0 & 25.0 & 0.0 & 0.0 & 0.0 & 2.0 & 0.0 & 2.0 & 58.0 & 2.0 & 57.0 \\
  & 78B & 43.2 & 0.0 & 43.2 & 2.4 & 0.0 & 2.4 & \textbf{6.0} & 0.0 & \textbf{6.0} & 73.0 & \textbf{6.0} & 70.0 \\
[-2ex]\\
 &  & \multicolumn{3}{c}{QD} & \multicolumn{3}{c}{CIT} & \multicolumn{3}{c}{COCO} & \multicolumn{3}{c}{avg.} \\
 family & model & \textit{agg} & \textit{cc} & \textit{ps} & \textit{agg} & \textit{cc} & \textit{ps} & \textit{agg} & \textit{cc} & \textit{ps} & \textit{agg} & \textit{cc} & \textit{ps} \\
\toprule
 LLaVA-1.5 & 7B & 39.0 & 3.0 & 36.0 & 4.0 & 2.0 & 3.0 & 4.0 & 3.0 & 1.0 & 10.2 & 1.9 & 9.0 \\
 \rowcolor{colorline} & Mistral-7B & 54.0 & 3.0 & 51.0 & 8.0 & 2.0 & 6.0 & 7.0 & 1.0 & 7.0 & 21.4 & 0.9 & 20.7 \\
 \rowcolor{colorline}\multirow{-2}{*}{LLaVA-NeXT} & Vicuna-7B & 70.0 & 2.0 & 68.0 & 7.0 & 0.0 & 7.0 & 8.0 & 1.0 & 7.0 & 24.5 & 0.4 & 24.1 \\
  & 0.5B & 11.0 & 1.0 & 10.0 & 2.0 & 1.0 & 1.0 & 2.0 & 2.0 & 1.0 & 5.9 & 0.9 & 5.2 \\
  LLaVA-OV & 7B & 41.0 & 1.0 & 41.0 & 2.0 & 1.0 & 1.0 & 3.0 & 0.0 & 3.0 & 15.1 & 0.4 & 15.0 \\
  & 72B & 53.0 & 2.0 & 51.0 & 6.0 & 1.0 & 5.0 & 11.0 & 3.0 & 9.0 & 24.0 & 1.0 & 23.2 \\
 \rowcolor{colorline}SmolVLM2 & 2.2B & 5.0 & 0.0 & 5.0 & 6.0 & 2.0 & 4.0 & 2.0 & 2.0 & 0.0 & 3.4 & 0.6 & 2.8 \\
 Idefics3 & 8B & 12.0 & 2.0 & 10.0 & 4.0 & 0.0 & 4.0 & 3.0 & 2.0 & 2.0 & 5.1 & 0.7 & 4.6 \\
 \rowcolor{colorline} & 3B & 27.0 & 0.0 & 27.0 & 5.0 & 0.0 & 5.0 & 4.0 & 3.0 & 1.0 & 10.1 & 0.6 & 9.5 \\
 \rowcolor{colorline} & 7B & 69.0 & 1.0 & 68.0 & 15.0 & 0.0 & 15.0 & 5.0 & 0.0 & 5.0 & 24.0 & 0.3 & 23.8 \\
 \rowcolor{colorline} & 32B & \textbf{84.0} & 1.0 & \textbf{83.0} & 11.0 & 1.0 & 11.0 & 11.0 & 1.0 & 10.0 & 31.4 & 0.4 & 31.1 \\
 \rowcolor{colorline}\multirow{-4}{*}{Qwen-2.5-VL} & 72B & 83.0 & 0.0 & \textbf{83.0} & \textbf{28.0} & 1.0 & \textbf{27.0} & 20.0 & 4.0 & 17.0 & \textbf{40.0} & 0.7 & \textbf{39.5} \\
  & 1B & 44.0 & \textbf{4.0} & 40.0 & 1.0 & 0.0 & 1.0 & 4.0 & 0.0 & 4.0 & 12.4 & 0.9 & 11.5 \\
  & 2B & 35.0 & 3.0 & 32.0 & 4.0 & 1.0 & 3.0 & 7.0 & 2.0 & 5.0 & 15.7 & 1.1 & 14.9 \\
 InternVL3 & 8B & 46.0 & 1.0 & 45.0 & 6.0 & 0.0 & 6.0 & 8.0 & 4.0 & 5.0 & 20.6 & 0.7 & 20.0 \\
  & 38B & 76.0 & 3.0 & 73.0 & 12.0 & \textbf{3.0} & 9.0 & 13.0 & 6.0 & 8.0 & 26.6 & 2.0 & 24.9 \\
  & 78B & \textbf{84.0} & 2.0 & 82.0 & 12.0 & 1.0 & 11.0 & \textbf{29.0} & \textbf{8.0} & \textbf{24.0} & 35.7 & \textbf{2.4} & 34.1 \\
    \end{tabularx}
    \label{tab:free_form_hint}
\end{table*}

\begin{table*}[t]
    \centering
    \scriptsize
    \caption{\textbf{Multi-turn OEG results on \bench{}.} Accuracy (\textit{acc}) and proactive suggestion rate (\textit{psr}) of MLLMs across all \bench{} splits.}
    \resizebox{\linewidth}{!}{
    \begin{tabular} { 
        rlcccccccccccccccc
    }
      &  & \multicolumn{2}{c}{ROD} & \multicolumn{2}{c}{VSOD} & \multicolumn{2}{c}{MVP-N} & \multicolumn{2}{c}{IN-C} & \multicolumn{2}{c}{QD} & \multicolumn{2}{c}{CIT} & \multicolumn{2}{c}{COCO} & \multicolumn{2}{c}{avg.} \\
family & model & \textit{acc} & \textit{psr} & \textit{acc} & \textit{psr} & \textit{acc} & \textit{psr} & \textit{acc} & \textit{psr} & \textit{acc} & \textit{psr} & \textit{acc} & \textit{psr} & \textit{acc} & \textit{psr} & \textit{acc} & \textit{psr}\\
\toprule
LLaVA-1.5 & 7B & \textbf{1.1} & 0.1 & 0.0 & 0.0 & 1.0 & 0.0 & 3.0 & 0.1 & 1.0 & 0.0 & \textbf{5.0} & 0.0 & \textbf{5.0} & 0.0 & 2.3 & 0.0 \\
\rowcolor{colorline} & Mistral-7B & \textbf{1.1} & 0.0 & \textbf{2.4} & 0.0 & 0.0 & 0.1 & 1.0 & 0.5 & 3.0 & 0.2 & 3.0 & 0.1 & 0.0 & 0.0 & 1.5 & 0.1 \\
\rowcolor{colorline}\multirow{-2}{*}{LLaVA-NeXT} & Vicuna-7B & 0.0 & 0.1 & \textbf{2.4} & 0.0 & 1.0 & 0.0 & 2.0 & 0.6 & \textbf{4.0} & 0.3 & 3.0 & 0.0 & 2.0 & 0.0 & 2.1 & 0.1 \\
 & 0.5B & 0.0 & 0.0 & 0.0 & 0.0 & 1.0 & 0.0 & 0.0 & 0.1 & 1.0 & 0.0 & 0.0 & 0.0 & 3.0 & 0.0 & 0.7 & 0.0 \\
 \multirow{-2}{*}{LLaVA-OV} & 7B & 0.0 & 0.0 & 0.0 & 0.0 & 1.0 & 0.0 & 0.0 & 0.4 & 2.0 & 0.0 & 1.0 & 0.0 & 2.0 & 0.0 & 0.9 & 0.1 \\
\rowcolor{colorline}SmolVLM2 & 2.2B & 0.0 & 0.0 & 0.0 & 0.0 & 2.0 & 0.0 & 2.0 & 0.0 & 1.0 & 0.0 & 0.0 & 0.0 & 2.0 & 0.0 & 1.0 & 0.0 \\
Idefics3 & 8B & 0.0 & 0.0 & 0.0 & 0.0 & 1.0 & 0.0 & 1.0 & 0.1 & 0.0 & 0.0 & 0.0 & 0.0 & 0.0 & 0.0 & 0.3 & 0.0 \\
\rowcolor{colorline} & 3B & 0.0 & 0.0 & 0.0 & 0.0 & 4.0 & 0.0 & 3.0 & 0.3 & 0.0 & 0.0 & 1.0 & 0.0 & 2.0 & 0.0 & 1.4 & 0.0 \\
\rowcolor{colorline}\multirow{-2}{*}{Qwen-2.5-VL} & 7B & \textbf{1.1} & 0.0 & \textbf{2.4} & 0.0 & \textbf{6.0} & 0.0 & \textbf{10.0} & 0.6 & 3.0 & 0.1 & 4.0 & 0.0 & 1.0 & 0.0 & \textbf{3.9} & 0.1 \\
 & 1B & 0.0 & 0.0 & \textbf{2.4} & 0.0 & 2.0 & 0.0 & 3.0 & 0.2 & 1.0 & 0.1 & \textbf{5.0} & 0.0 & 3.0 & 0.0 & 2.3 & 0.1 \\
InternVL3 & 2B & 0.0 & 0.0 & 0.0 & 0.0 & 3.0 & 0.0 & 2.0 & 0.2 & 0.0 & 0.0 & 0.0 & 0.0 & 2.0 & 0.0 & 1.0 & 0.0 \\
 & 8B & 0.0 & 0.0 & 0.0 & 0.0 & 5.0 & 0.0 & 2.0 & 0.2 & 1.0 & 0.0 & 2.0 & 0.0 & 1.0 & 0.0 & 1.6 & 0.0 \\
    \end{tabular}
    }
    \label{tab:free_form_multi_turn}
\end{table*}

\section{Broader impacts statement}
\label[appendix]{appx:broader_impact}
\bench{} is designed to assess the proactiveness of multimodal large language models (MLLMs), \ie, their ability to request additional input when faced with ambiguous or insufficient visual information. 
As MLLMs are increasingly deployed in interactive and safety-critical applications (\ie, assistive tools, autonomous systems), encouraging and evaluating such behavior is essential for developing more collaborative and user-aligned AI.
By highlighting current models' proactiveness limitations, our work provides meaningful insights for researchers seeking to build more collaborative AI systems.
However, promoting proactiveness must be carefully balanced to avoid over-questioning or inefficient behavior. 
While our benchmark promotes interpretability and safe failure modes (\ie, abstention over hallucination), there is a risk of misuse in adversarial settings if models over-rely on user feedback. 
We release \bench{} to support reproducible and community-driven progress toward more robust and human-aware MLLMs.

\section{Licenses}
\label[appendix]{appx:license}
All original material presented in this work is intended solely for academic research and not for commercial purposes. 
Below, we report the licenses of the used datasets and models:
\begin{itemize}[leftmargin=10pt]
    \item ROD~\cite{lee2023hardwiring}: This dataset is released without a license.
    \item VSOD~\cite{liao2020occlusion}: MIT License.
    \item MVP-N~\cite{wang2022mvp}: MIT License.
    \item ImageNet-C~\cite{hendrycks2019benchmarking}: Apache License 2.0.
    \item QuickDraw~\cite{quickdraw}: CC-BY-4.0.
    \item ChangeIt~\cite{soucek2022lookforthechange}: MIT License.
    \item MS-COCO~\cite{lin2014microsoft}: CC-BY-4.0.
    \item \llavacvpr{}~\cite{liu2024improved}: Llama2.
    \item \llavanext{} Vicuna~\cite{liu2024improved}: Llama2.
    \item \llavanext{} Mistral~\cite{liu2024improved}: Apache License 2.0.
    \item \llavaov{}~\cite{li2024llava}: Apache License 2.0.
    \item \qwen{}~\cite{bai2025qwen2}: Apache License 2.0.
    \item \smolvlm{}~\cite{marafioti2025smolvlm}: Apache License 2.0.
    \item \idefics{}~\cite{laurenccon2024building}: Apache License 2.0.
    \item \internvl{}~\cite{zhu2025internvl3}: Apache License 2.0.
    \item \instructblip{}~\cite{instructblip}: Llama2.
    \item \phimultimodal{}~\cite{abouelenin2025phi}: MIT License.
\end{itemize}

\section{LLM usage declaration}
\label[appendix]{appx:llm_usage}
During the writing of this paper, we used LLMs for polishing writing and proofreading the manuscript.

\end{document}